\documentclass{article}

\usepackage{arxiv}

\usepackage[utf8]{inputenc} % allow utf-8 input
\usepackage[T1]{fontenc}    % use 8-bit T1 fonts
\usepackage{hyperref}       % hyperlinks
\usepackage{url}            % simple URL typesetting
\usepackage{booktabs}       % professional-quality tables
\usepackage{amsfonts}       % blackboard math symbols
\usepackage{nicefrac}       % compact symbols for 1/2, etc.
\usepackage{microtype}      % microtypography
\usepackage{lipsum}		% Can be removed after putting your text content
\usepackage{newunicodechar}
\newunicodechar{−}{-}
\usepackage{threeparttable} % for threeparttable environment
\usepackage{graphicx}
\usepackage{natbib}
\usepackage{doi}

\title{EduArt: An educational-level benchmark for evaluating art history knowledge in large language models\thanks{This is a preprint. The manuscript is currently under review at \textit{Computational Humanities Research} (CHR), submitted to the Call for Papers ``Computational Approaches to Art.''}}

\date{} 					% Or removing it

\author{ \href{https://orcid.org/0000-0002-3504-3241}{\includegraphics[scale=0.06]{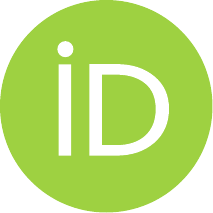}\hspace{1mm}Gianmarco Spinaci}\\
    Department of Classical Philology and Italian Studies, University of Bologna, Bologna, Italy\\
    Villa i Tatti -- The Harvard University Center for Italian Renaissance Studies, Firenze, Italy\\
    \texttt{gianmarco.spinaci2@unibo.it}\\
	\And
	\href{https://orcid.org/0000-0002-9620-7107}{\includegraphics[scale=0.06]{orcid.pdf}\hspace{1mm}Lukas Klic} \\
    Villa i Tatti -- The Harvard University Center for Italian Renaissance Studies, Firenze, Italy\\
	\texttt{lklic@itatti.harvard.edu} \\
    \And
	\href{https://orcid.org/0000-0002-9806-084X}{\includegraphics[scale=0.06]{orcid.pdf}\hspace{1mm}Giovanni Colavizza} \\
    Department of Classical Philology and Italian Studies, University of Bologna, Bologna, Italy\\
    Center for Computational and Digital Humanities, University of Copenhagen, Copenhagen, Denmark\\
	\texttt{giovanni.colavizza@unibo.it} \\
}

\renewcommand{\headeright}{}
\renewcommand{\undertitle}{}
\renewcommand{\shorttitle}{\textit{arXiv} Template}

\hypersetup{
pdftitle={A template for the arxiv style},
pdfsubject={q-bio.NC, q-bio.QM},
pdfauthor={David S.~Hippocampus, Elias D.~Striatum},
pdfkeywords={First keyword, Second keyword, More},
}

\begin{document}
\maketitle

\begin{abstract}
Large language models are now able to score ceiling performance on general benchmarks, but these aggregate measures reveal little about how models behave within single disciplines. Moreover, existing art-focused evaluations have relied on synthetically generated questions and have rarely reported item-level properties.
 
This paper introduces EduArt, an educational-level benchmark for evaluating art-historical knowledge and visual reasoning in multimodal LLMs. EduArt comprised 871 human-authored questions drawn from Italian secondary-school exercises and the United States Advanced Placement Art History examinations, spanning two languages and seven question formats that ranged from multiple choice to in-text word placement and error identification. Twelve models from six provider families were evaluated under two conditions, a default answer-only condition and a motivation condition requiring a written justification, and the benchmark was characterized using Classical Test Theory and a logistic regression isolating the effects of format, language, image presence, and model.
 
The benchmark showed strong psychometric properties (mean discrimination 0.514, with 82.3 percent of good discriminator items), while multiple-choice accuracy saturated near ceiling for six models, confirming that recognition formats alone are not enough to distinguish frontier models. The question format was a strong independent predictor of accuracy: models exceeding 94 percent on multiple choice can fall to 23.9 percent on open completion (Claude Opus 4.6) and 6.2 percent on error identification (Claude Sonnet 4.6).
 
The motivation condition changed accuracy in a predominantly negative and model-family-dependent direction. These dissociations indicated that art-historical knowledge and the ability to deploy it were distinct capabilities, and that single-format benchmarks systematically overestimated what models could reliably do. Mapping this fine-grained capability profile was found to be a precondition for the responsible use of multimodal LLMs in art-historical scholarship, where tasks demand the production and manipulation of content rather than selection from fixed options.
\end{abstract}

% keywords can be removed

\section{Introduction}
Over the past years, large language models (LLMs) have made remarkable progress on academic knowledge benchmarks. On the Massive Multitask Language Understanding (MMLU) benchmark \citep{hendrycks_measuring_2021}, which spans 57 subjects among STEM and Humanities subjects, recent models achieved around 90 percent accuracy \citep{openai_gpt-4_2024, team_gemini_2025, claude_nodate}. As these tasks become saturated, more challenging benchmarks emerge. MMLU-Pro raised the difficulty by adding options, focusing on reasoning, and removing trivial questions \citep{wang_mmlu-pro_2024}. BIG-Bench collected over 200 tasks from 450 contributors, covering biology, physics, math, and social biases, designed to challenge LLMs \citep{srivastava_beyond_2023}. GPQA presents 448 domain-expert-written questions in biology, physics, and chemistry, tested by non-experts and LLMs \citep{rein_gpqa_2024}. Humanity's Last Exam comprises 2,500 expert-crafted questions across hundreds of subjects, aiming to serve as the final closed-ended academic benchmark \citep{phan_humanitys_2026}. Yet, even this effort yields roughly 25 questions per topic, resulting in an underrepresentation of elements for diagnosing strengths and weaknesses within single disciplines. These general-purpose benchmarks have been instrumental in tracking aggregate models' progress, but they share a structural trade-off: broad coverage across various domains at the expense of diagnostic depth within any single one \citep{chang_survey_2024}.
 
Building on these broad evaluations, the research community has produced more in-depth analyses. These analyses are particularly concentrated in domains with significant economic and societal impact. For example, in the medical field, studies demonstrated that ChatGPT's performance on the United States Medical Licensing Exam (USMLE) is around 60 percent without specialized training \citep{kung_performance_2023}, with specific Deepseek configurations substantially exceedeing this level, by reaching 93 percent accuracy \citep{siam_benchmarking_2025}. In law, GPT-4 passed the bar exam, the professional licensure exam required to practice law in the USA \citep{katz_gpt-4_2024}. LegalBench provided a legal reasoning benchmark consisting of 162 tasks \citep{guha_legalbench_2023}. In science education, ScienceQA introduced multimodal questions with educational framing via "Thought Chains" \citep{lu_learn_2022}. One thing stands out across these studies: when researchers use real-world materials from specific fields, they can uncover the unique strengths and weaknesses of language models in that discipline. Broader, more general tests often miss these insights.
 
\subsection{Multimodality and Visual Reasoning}
 
In parallel with domain-specific evaluations, the emergence of Vision-Language and multimodal models required expanding the scope of LLM evaluation. These models can be tested beyond text-only questions, opening the door to tasks that require joint reasoning across images and language. Visual Question Answering (VQA) became the standard evaluation framework, where the examined model is presented with an image and a natural-language question and must produce an answer that draws on both modalities. Early VQA benchmarks, e.g., VQAv2 \citep{goyal_making_2017}, focused on generic visual understanding, such as recognizing objects, spatial relationships, and colors in everyday photographs. Recent efforts have raised the level considerably. MMMU introduced college-level, multidisciplinary questions requiring multimodal reasoning \citep{yue_mmmu_2024}. Culturally diverse benchmarks like CVQA \citep{romero_cvqa_2024} have highlighted that most VQA datasets are biased towards English and Western content, and that models exhibit significant performance degradation across different scenarios.
 
These developments introduce a more fundamental question: what does it mean for a generative model to perform visual reasoning? The literature shows several angles of approach. Pre-generative LLMs' foundational works framed the evaluation of visual reasoning capabilities through the combination of the model's perception and multi-step operations such as counting, comparing, and spatial relation \citep{johnson_clevr_2017}. Another approach, formalized as Visual Commonsense Reasoning, characterized visual reasoning as inferences beyond the only recognition, requiring world knowledge grounded in the extracted visual evidence \citep{zellers_recognition_2019}. More recent approaches have increasingly analyzed the task, highlighting the distinction between apparent and genuine visual reasoning. With the benchmark MathVerse \citep{zhang_mathverse_2025}, the authors tested a set of questions supported with images of mathematical diagrams. They observed that as they progressively removed visual information from the questions, model performance often did not degrade as expected, suggesting that generative multimodal LLMs are often prone to relying on standard language reasoning, which can incorrectly attribute language-driven results to visual reasoning. Another study \citep{tong_eyes_2024} documented spatial relations, fine-grained discrimination, and counting as failure modes persistent in small and bigger multimodal models, despite the parameter scale, suggesting that these failures reflect architectural limits rather than data deficits. Together, the literature establishes that high accuracy on multimodal benchmarks does not, on its own, demonstrate visual reasoning, and that distinguishing genuine visual reasoning from language-shortcut answering requires evaluation designs that constrain or expose the shortcut path.
 
\subsection{Visual Reasoning Evaluation in Cultural Heritage}
 
These methodological developments have established VQA as a mature evaluation paradigm, but they have not been applied extensively to cultural heritage applications. Starting with its limited application in specialized domains, efforts in cultural heritage have begun to explore VQA for artworks. AQUA \citep{garcia_dataset_2020} was the first approach, with question-answer pairs automatically generated from paintings and associated textual descriptions, followed by a crowdsourced cleaning process. VISCOUNTH introduced a large-scale multilingual dataset for cultural heritage \citep{becattini_viscounth_2023}. Most recently, VQArt-Bench \citep{alfarano_vqart-bench_2025} proposed a semantically rich benchmark constructed through a multi-agent pipeline, explicitly noting that prior benchmarks' shallow syntactic structures incentivized models to exploit statistical shortcuts. Other works have addressed narrower visual tasks. For example, \citep{spinaci_benchmarking_2025} benchmarked LLMs and vision-language models on single-label classification of Christian saints, finding that frontier models can outperform supervised baselines on curated iconographic datasets.
 
Despite these exploratory efforts in art-focused VQA, there remain two important limits on their art-historical applicability. First, the questions in these benchmarks are predominantly generated synthetically. Second, they are designed to benchmark general-purpose model capabilities on VQA tasks, rather than to diagnose art-historical knowledge. In other words, the existing landscape offers either narrow, task-specific evaluations or broader question-answering with synthetically generated content; not yet an educationally grounded assessment within a single art-historical evaluation framework. Moreover, existing benchmarks do not report item-level properties, making it impossible to determine whether observed model failures reflect genuine knowledge gaps, task format sensitivity, or poorly constructed items.
 
\subsection{Educational materials and Classical Test Theory}
 
Art History presents a particularly interesting case for model evaluation, as it combines several challenges that general benchmarks do not fully isolate. First, it is an inherently multimodal field, as understanding and analyzing artworks requires detecting visual features (composition, color, techniques, etc). Second, it spans different cognitive levels, from straightforward factual recall (identifying style or technique) to interpretive reasoning (reading symbolic meaning from visual evidence). And third, it raises a distinctive epistemic challenge. Educational-level art history questions have agreed-upon correct answers verified by professional educators and examination boards. Expert-level questions often become genuinely contested, with attributions debated rather than converging. An educational-level benchmark thus provides verifiable ground truth while still testing a meaningful range of knowledge and reasoning.
 
Beyond aggregate accuracy, recent work has begun to apply psychometric frameworks from educational measurement to LLM benchmarks, such as Classical Test Theory (CTT), which characterizes questions in terms of two dimensions. One is the difficulty, which is the proportion of respondents answering correctly. The second is the discrimination as a point-biserial correlation between item-level performance and total score. CTT provides a way to diagnose which item meaningfully distinguishes strong from weak models. Item Response Theory (IRT) extends the previous framework by modeling the probability of a correct response as a function of examinee ability and item parameters, including guessing. It has been applied to demonstrate that even a small subset of carefully selected items from benchmarks such as MMLU can still recover aggregate performance similar to the source, proving that most items only contribute a little diagnostic signal once a benchmark saturates \citep{polo_tinybenchmarks_2024}. IRT has also been applied to provide a granular view of benchmark validity beyond accuracy, by characterizing items and model families \citep{zhou_lost_2026}.
 
Taking these considerations into account, we propose EduArt, an educational-level art historical benchmark created with the need to answer the following research questions: \textbf{RQ1:} Does an educational-level benchmark constitute a valid tool for evaluating multimodal LLMs on art-historical knowledge and visual reasoning, and what do item-level characteristics reveal about the sources of difficulty? \textbf{RQ2:} To what extent does question format independently affect measured performance? In other words, do models that demonstrate strong art-historical knowledge on multiple-choice items maintain that performance when tested through alternative formats such as true/false statements, correct word in-text placement, and error identification? \textbf{RQ3:} How does requiring models to provide a motivation for their answer affect accuracy, and does this effect differ systematically across model families and question types?

\section{Methodology}

This section describes the methodology adopted to address the three research questions introduced above. It covers the construction and composition of the benchmark dataset, the extraction pipeline used to collect and normalize questions from heterogeneous sources, the selection of models and evaluation framework, and the metrics used to assess performance.

\subsection{The questions} 

The benchmark dataset was constructed from two independent sources that cover complementary linguistic and educational content. 

The first source is the MyZanichelli digital exercise platform, published by Zanichelli, a major Italian academic publisher. The platform organizes content into numbered exercises, each holding several questions grouped by theme (e.g., \textit{Art History} and \textit{Mathematics}) and with various formats, see Table \ref{tab:question_types}. An important feature of these questions resides in their human-made nature. Professional educators designed questions to accompany art history textbooks in Italian secondary education. For the dataset, we restricted the collection to questions on the only theme of \textit{Pittura rinascimentale} (renaissance paintings), yielding 57 exercises and 668 questions in Italian\footnote{The exercises can be found online at the  \href{https://esercizi.zanichelli.it/argomento/Pittura-rinascimentale/x2rnbu-x2rnih-hv05y-2n\_5c}{portal website}. The access is login-protected.}.

The second source is a set of released examination papers from the College Board Advanced Placement (AP) Art History program, covering the years 2013-2024. These are standardized, nationally administered examinations taken by high-school students in the United States as part of a college-level art history curriculum. This source contributed 203 questions in English, all in a single-answer multiple-choice format.

Together, these sources ensure both linguistic and format diversity while maintaining a consistent pedagogical foundation. All questions are designed to assess learning objectives defined by educational programs, providing a reliable ground truth without requiring the generation of synthetic items and avoid pedagogical assumptions.

The questions used in this benchmark are educational materials protected by copyright. Therefore, the set cannot be redistributed alongside the paper. To support transparency and reproducibility, the public repository contains the code required to process the inputs and reconstruct the evaluation pipeline, while the extraction procedure from the original sources is described in the following section. 

\subsubsection{Extraction Pipeline}

Because of the sources' different formats, the process for extracting the questions into a usable version for the benchmark followed two procedures. Questions from MyZanichelli were collected through automated browser interaction using the Playwright tool\footnote{\url{https://playwright.dev/}}. For each identified question, the script randomly selected or wrote the answer based on the expected format,  submitted the question, captured a full-width screenshot of the window, and extracted the resulting HTML. The resulting screenshot and source page include visual cues that indicate correct or incorrect answers. The question text and structured answer choices were not directly available in the HTML source, as the platform uses a dynamic framework for populating the pages. Therefore, the final version is the structured output of an extraction process that submits each question screenshot and its text to generative models. To mitigate errors in this phase, input files were processed independently by two different models (Google Gemini 3 Flash and Anthropic Sonnet 4.5), and the outputs were compared. Discrepancies between the two extractions were flagged for manual review. After resolving disagreements and refining the extraction prompt based on observed error patterns, the final extraction was performed with a second run, using the same two models, followed by another manual review phase. At the final step, the discrepancies fell in only two categories, accents and double spaces, related to the structure of the answer rather than its content, proving the prompt effectiveness\footnote{The extraction prompt is available online at \url{https://github.com/llm-art/EduArt/tree/main/prompts}}. For AP Art History, the released examination papers, available as PDF documents, were parsed programmatically. Questions, answer options, and correct answers were extracted and structured into the same schema used for the Zanichelli material. 

Finally, each question, regardless of source, was normalized into a canonical JSON record containing the question text, an ordered list of answer options with alphabetic keys, the ground-truth answer, a flag indicating the presence of an associated artwork image, the image file path, the language code (Italian or English), and an exercise or section title where available.

\subsubsection{Dataset}

The dataset covers seven question formats of the following types: 

\begin{enumerate}
    \item \verb|multiple_choice_radio|
    \item \verb|multiple_choice_check|
    \item \verb|true_false|
    \item \verb|completion_closed|
    \item \verb|completion_open|
    \item \verb|positioning|
    \item \verb|select_errors|
\end{enumerate}

\textit{Multiple\_choice\_radio} requires a single correct answer from a set of options,
 \textit{multiple\_choice\_check}, multiple correct answers from a set of options, \textit{true\_false} a series of statement with true or false option, \textit{positioning}, drag-and-drop the correct word among texts, \textit{completion\_closed} fill-in-the-blank with a constrained word bank, \textit{completion\_open}, fill-in-the-blank without any constraint, and \textit{select\_errors}, identification of incorrect words in a text. The total number of questions per type is available in Tab \ref{tab:question_types}. Multiple choice questions will be addressed as MCQ throughout the paper.

\begin{table}[hbt!]
\centering
\caption{Distribution of question types across the EduArt benchmark. Source: Authors elaboration from MyZanichelli (Zanichelli) and College Board AP Art History released examinations (2013–2024).}
\begin{threeparttable}
\label{tab:question_types}
\begin{tabular}{ll}
\toprule
Type & Count \\
\midrule
\texttt{multiple\_choice\_radio} & 370 \\
\texttt{multiple\_choice\_check} & 117 \\
\texttt{true\_false} & 83 \\
\texttt{positioning} & 108 \\
\texttt{completion\_closed} & 69 \\
\texttt{completion\_open} & 75 \\
\texttt{select\_errors} & 49 \\
\midrule
\textbf{Total} &  \textbf{871} \\
\bottomrule
\end{tabular}
\end{threeparttable}
\end{table}

In addition to types, the datasets is analyzed across textual, visual, and structural domains. Textual features include word count, token count, number of answer options, and the average length of those options. Word counts are computed with spaCy language-specific tokenizers\footnote{including Italian (\textbf{it\_core\_news\_sm}) and English (\textbf{en\_core\_web\_sm})}, and token counts are computed using the \textit{o200k\_base} encoder. The statistics presented in  Tab \ref{tab:question_length_stats}, include title, instruction, text, and options where available.

\begin{table}[hbt!]
\centering
\caption{Descriptive statistics of question length across all 871 items, including title, instruction, body text, and answer options where present. Word counts use spaCy language-specific tokenizers; token counts use the o200k\_base encoder. Source: Authors' computation.}
\begin{threeparttable}
\label{tab:question_length_stats}
\begin{tabular}{lrrrrr}
\toprule
 Metric & Mean & Median & Std & Min & Max \\
\midrule
 Word count & 53.4 & 54.0 & 32.9 & 4 & 148 \\
 Token count & 91.2 & 92.0 & 59.0 & 8 & 250 \\
\bottomrule
\end{tabular}
\end{threeparttable}
\end{table}

The dataset comprises 261 distinct images out of 436 questions flagged as having images. This divergence arises because multiple AP questions share a single image. Nevertheless, each question is treated singularly. All images are low to medium resolution images, with mean dimensions of approximately 470 x 463 pixels and a mean size of 213.3 KB, see Tab. \ref{tab:image_stats}. Resolution range is sufficient for detecting broad compositional features, as it places within the range commonly used for web-based visual content. It may be insufficient for tasks requiring fine-grained visual discrimination such as distinguish brushwork technique or identifying secondary figures in background. For this reason, results on image-dependent questions should be interpreted as a lower bound of what models could achieve with higher-fidelity reproductions.

\begin{table}[hbt!]
\centering
\caption{Width, height, and file-size statistics for the 261 distinct images associated with image-flagged items. Source: Authors' computation.}
\label{tab:image_stats}
\begin{tabular}{lrrrr}
\toprule
\textbf{Metric} & \textbf{Mean} & \textbf{Std} & \textbf{Min} & \textbf{Max} \\
\midrule
Width (px) & 470 & 297 & 180 & 1568 \\
Height (px) & 463 & 193 & 194 & 1568 \\
File size (KB) & 213.3 & 55.4 & 61.5 & 435.0 \\
\bottomrule
\end{tabular}
\end{table}

\subsection{Evaluation framework}

\begin{table*}[t]
\centering
\caption{Models and their exact API identifiers or dated model versions used in the benchmark, organised by provider family. Gemini models, last accessed March 2026. Source: Authors' compilation from OpenAI, Google AI Studio, Anthropic, and Amazon Web Services Bedrock model documentation, accessed February–May 2026}
\label{tab:model_versions}
\begin{tabular}{ll}
\toprule
\textbf{Name} & \textbf{Version / API identifier} \\
\midrule
\textit{GPT-5.5} & \texttt{gpt-5.5-2026-04-23} \\
\textit{GPT-5.4 Mini} & \texttt{gpt-5.4-mini-2026-03-17} \\
\textit{GPT-5.4 Nano} & \texttt{gpt-5.4-nano-2026-03-17} \\

\textit{Gemini 3.1 Pro Preview*} & \texttt{gemini-3.1-pro} \\
\textit{Gemini 3.1 Flash Lite Preview*} & \texttt{gemini-3.1-flash-lite-preview}  \\

\textit{Gemini 3.5 Flash*} & \texttt{gemini-3.5-flash} \\

\textit{Claude Opus 4.6} & \texttt{us.anthropic.claude-opus-4-6-v1} \\
\textit{Claude Sonnet 4.6} & \texttt{us.anthropic.claude-sonnet-4-6} \\
\textit{Claude Haiku 4.5} & \texttt{us.anthropic.claude-haiku-4-5-20251001-v1:0} \\
\textit{Qwen3-VL-235B} & \texttt{qwen.qwen3-vl-235b-a22b} \\
\textit{Mistral Large 3 675B} & \texttt{mistral.mistral-large-3-675b-instruct} \\
\textit{Pixtral Large} & \texttt{us.mistral.pixtral-large-2502-v1:0} \\
\bottomrule
\end{tabular}
\end{table*}

Twelve models spanning 6 provider families has been evaluated and selected to represent the current range of multimodal LLM capabilities: frontier reasoning models, mid-tier, their lightweight alternatives, and open-weight architectures. This selection enables comparison across provider families (OpenAI, Google, Anthropic, Qwen, Mistral, Meta), within families across capability tiers (e.g., GPT-5.5 vs. GPT-5.4 Mini and Nano), and between proprietary and open-weight models. All Google models were queried via the Gemini API and OpenAI models via the OpenAI API. all other models were accessed through Amazon Web Services (AWS) Bedrock. Table \ref{tab:model_versions} lists the models and their identifiers. For reproducibility, the table reports the exact API identifiers or dated model versions used in the benchmark. For \textit{Gemini} family, where public model names may remain stable despite backend updates, the models were accessed throughout March 2026. \textit{Gemini 3.1 Flash Lite Preview} last updated was on March 2026, \textit{Gemini 3.5 Flash} on May 2026, and \textit{Gemini 3.1 Pro Preview} on February 2026.

Each question was submitted to the model a system prompt and a user prompt. The system prompt is fixed and frames the task as an art history examination. The model is instructed to always provide an answer and to return its response in a strict JSON schema. When uncertain, the model is directed to make an educated guess based on visual analysis, stylistic reasoning, and iconographic interpretation. It is followed by a user prompt containing the plain-text content of the question file. For questions with associated images, the image is passed as a base64-encoded attachment via the provider's vision interface.

Each model was evaluated under two conditions. In the default condition, the model is prompted to return only the answer in the required JSON schema. In the motivation condition, the model is additionally required to complete a mandatory reasoning field of 2–4 sentences, explicitly grounding the justification in art-historical knowledge, visual evidence, or contextual clues. The two conditions are applied to the same 871 items for each model, enabling a paired comparison of how elicited reasoning affects answer accuracy. The prompt are available in the online repository, see Data Availability section for the link.

\subsubsection{Temperature and Reproducibility}
Due to differences in provider-recommended settings and API constraints, temperature was configured per provider rather than uniformly across all models. For Anthropic Claude models, temperature was set to 0 to obtain near-deterministic outputs. For open-weight models accessed via AWS Bedrock, temperature was set to 0 following standard greedy decoding practice. For Google Gemini models, temperature was set to 1 with a seed where the API permitted, as lower values have been reported to degrade output quality for this model family. GPT-5.x models do not expose temperature as a configurable parameter and were run at the provider default of 1. Each combination of question, model, and condition was evaluated in a single forward pass. 

\subsubsection{Evaluation Metrics}

Evaluation metrics are stratified by question type to reflect the heterogeneity of answer structures. For \textit{multiple\_choice\_radio} items, performance is measured using exact match. It is 1 if the predicted option key matches the ground truth and 0 otherwise. For \textit{multiple\_choice\_check} items, the F1 score is computed over the set of selected options. \textit{True\_false} uses statement accuracy, corresponding to the proportion of individual statements correctly classified. For \textit{positioning} tasks, performance is measured using element accuracy, defined as the proportion of elements correctly placed. For completion tasks, both \textit{completion\_closed} and \textit{completion\_open}, blank accuracy is computed, defined as the proportion of white spaces (the blanks) correctly filled. For open completion, matching is performed using fuzzy string comparison with case and accent normalization, while close completion relies on normalized exact string matching. Finally, for \textit{select\_errors} items, the F1 score is computed over the predicted versus ground-truth sets.

The AP source contributes exclusively English-language questions in multiple-choice radio format. MyZanichelli contributes to different types of questions written in Italian other than MCQs. In other words, the dataset contains Italian and English MCQs, providing a direct basis for estimate a language effect under this format condition. For non-MCQs, language and format cannot be effectively separated, as they are all in Italian. A logistic regression model is fitted to the full dataset to estimate the independent contribution of question type, language, image presence, and model identity to the probability of a correct response, allowing these factors to be assessed simultaneously while controlling for one another. For RQ2, which examines format as an independent variable, the macro-average across all seven question types serves as the primary aggregate, assigning equal weight to each format regardless of its frequency. The MCQ exact match score is additionally reported as a format-controlled reference point, given its direct comparability across all models and its freedom from the source-level confound.

\section{Results}

This section presents the empirical findigs of the EduArt benchmark addressing the three research questions listed in the introduction. Results are reported under two experimental conditions: default, in which the model is prompted for the answer only, and motivation, in which the model is additionally asked to provide a brief motivation. Unless stated otherwise, primary results refer to the default condition.

\begin{table}[hbt!]
\centering
\caption{Macro-average accuracy across the seven question formats and exact-match accuracy on multiple-choice items (MCQ radio) with 95 percent binomial confidence intervals, under the default (answer-only) condition. Source: Authors' computation.}
\label{tab:model_ranking}
\begin{tabular}{l r r r}
\toprule
\textbf{Model} & \textbf{Macro Avg} & \textbf{$\mathrm{MCQ}$} & \textbf{95 \% CI ($\mathrm{MCQ}$)} \\
\midrule
\textit{Gemini 3.1 Pro Preview}        & \textbf{82.8 \%} & 95.7 \% & [93.1 \%, 97.3 \%] \\
\textit{GPT-5.5}                       & \textbf{80.0 \%} & 95.1 \% & [92.4 \%, 96.9 \%] \\
\textit{Gemini 3.5 Flash}              & \textbf{68.9 \%} & 96.5 \% & [94.1 \%, 97.9 \%] \\
\textit{Claude Opus 4.6}               & \textbf{67.3 \%} & 94.6 \% & [91.8 \%, 96.5 \%] \\
\textit{Claude Sonnet 4.6}             & \textbf{64.8 \%} & 94.3 \% & [91.5 \%, 96.3 \%] \\
\textit{Qwen3-VL-235B}                 & \textbf{64.5 \%} & 88.6 \% & [85.0 \%, 91.5 \%] \\

\textit{Gemini 3.1 Flash Lite Preview} & \textbf{62.8 \%} & 92.4 \% & [89.3 \%, 94.7 \%] \\

\textit{Mistral Large 3 675B}          & \textbf{58.8 \%} & 84.6 \% & [80.6 \%, 87.9 \%] \\
\textit{GPT-5.4 Mini}                  & \textbf{54.9 \%} & 82.7 \% & [78.5 \%, 86.2 \%] \\
\textit{Claude Haiku 4.5}              & \textbf{53.1 \%} & 84.0 \% & [80.0 \%, 87.4 \%] \\
\textit{GPT-5.4 Nano}                  & \textbf{40.1 \%} & 62.7 \% & [57.7 \%, 67.5 \%] \\
\textit{Pixtral Large}                 & \textbf{29.1 \%} & 28.6 \% & [24.3 \%, 33.5 \%] \\
\bottomrule
\end{tabular}
\end{table}

Table \ref{tab:model_ranking} reports the ranking of the evaluated models on macro-average across the seven question formats, and the MCQ exact-match score with 95 percent confidence interval (CI). Macro-average accuracy ranges from 29.1 percent (Pixtral Large) to 82.8 percent (Gemini 3.1 Pro Preview). 

For the MCQ exact-match, six models exceed 90 percent accuracy. In the same question type, scores range from 28.6 percent to 96.5 percent. The CI of the five best-performing models (Gemini 3.1 Pro Preview, Gemini 3.5 Flash, GPT 5.5, Claude Opus 4.6, and Claude Sonnet 4.6) overlap, with widths ranging from 3.8 to 4.8 percentage points.

\subsection{Item-level psychometric properties}

\begin{table}[hbt!]
\centering
\caption{Distribution of items across difficulty categories defined by the proportion of models answering each item correctly (p). Source: Authors' computation following the Classical Test Theory.}
\label{tab:item_difficulty}
\begin{tabular}{l c r}
\toprule
\textbf{Category} & \textbf{$p$ Range} & \textbf{Count} \\
\midrule
Very Easy & $\geq 0.90$    & 195 \\
Easy      & $0.70$--$0.89$ & 316 \\
Medium    & $0.30$--$0.69$ & 295 \\
Hard      & $0.10$--$0.29$ &  40 \\
Very Hard & $< 0.10$       &  25 \\
\midrule
\textbf{Total} & & \textbf{871} \\
\bottomrule
\end{tabular}
\end{table}

Table \ref{tab:item_difficulty} shows the full range of difficulty categories of the 871 items in the benchmark. There are 195 items classified as very easy (p $\geq$ 0.90), 316 as easy, 295 as medium, 40 as hard, and 25 as very hard (p < 0.10). Table \ref{tab:item_discrimination} reports the item discrimination scores, measured as the point-biserial correlation ($r_{pb}$) between item-level performance and total score. A high score indicates that models answering an item correctly tend to score well on the benchmark overall. Discrimination is computed on 831 of the 871 items. The remaining 40 items, including 37 MCQ radio items, were answered correctly by all models, showing zero variance across examinees and are therefore non-estimable.. MCQ radio type has the highest mean discrimination with a value of 0.608, followd by completion open (0.532), Positioning (0.505), and completion closed (0.501). At the lower end, true/false shows a mean discrimination of 0.245, with 25 and 24 items in the fair and poor range respectively.

\begin{table}[hbt!]
\centering
\caption{Item discrimination by question type (CTT). Good items have $r_{pb} \geq .30$, fair items $.10 \leq r_{pb} < .30$, and poor items $r_{pb} < .10$. 
Source: Authors' computation.}
\label{tab:item_discrimination}
\begin{tabular}{l r r r r r}
\toprule
\textbf{Type} & \textbf{N} & \textbf{Mean $r_{pb}$} & \textbf{Good} & \textbf{Fair} & \textbf{Poor} \\
\midrule
MCQ radio         & 333 & 0.608 & 305 & 11 & 17 \\
Completion open   &  75 & 0.532 &  67 &  6 &  2 \\
Positioning       & 106 & 0.505 &  84 & 16 &  6 \\
Completion closed &  69 & 0.501 &  58 &  9 &  2 \\
MCQ check         & 117 & 0.471 &  91 & 18 &  8 \\
Select errors     &  48 & 0.445 &  45 &  3 &  0 \\
True/false        &  83 & 0.245 &  34 & 25 & 24 \\
\midrule
All items         & 831 & 0.514 & 684 & 88 & 59 \\
\bottomrule
\end{tabular}
\end{table}

\begin{figure*}
    \centering
    \includegraphics[width=1\linewidth]{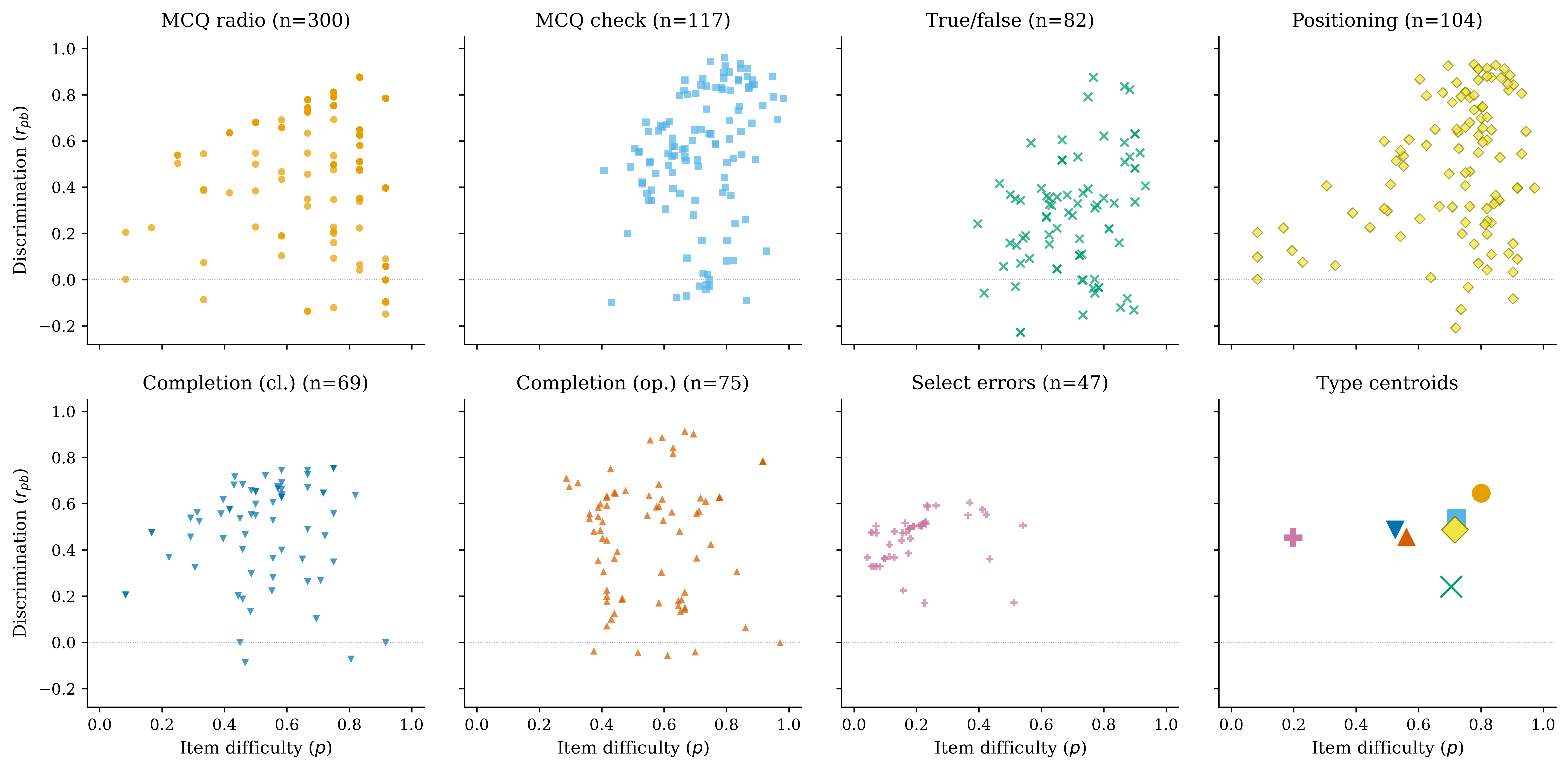}
    \caption{Item difficulty ($p$) plotted against discrimination ($r_{pb}$) for
each of the 871 benchmark items, grouped by question type.Horizontal reference lines mark the discrimination thresholds for good ($r_{pb} = 0.30$) and fair ($r_{pb} = 0.10$) items, and the vertical line marks the mean item difficulty. The final panel shows the per-type centroids. The number of items with zero variance across examinees are excluded.}
    \label{fig:difficulty}
\end{figure*}

Figure \ref{fig:difficulty} plots difficulty ($p$) against discrimination ($r_{pb}$) for each item grouped by question type. The last panel summarizes the centroids and highlights the clustering of MCQ radio and MCQ check around high mean difficulty ($p = 0.84$ and $p = 0.73$ respectively) with medium-high discrimination. Positioning ($p = 0.71$) positions near this group on the difficulty axis, but finds a slightly higher discrimination ($r_{pb}=0.51$). Completion Open ($p=0.57$) and completion closed ($p=0.55$) form a middle group. The true/false type shows moderate difficulty with the lowest discrimination of all formats ($r_{pb}=0.25$). Select errors is isolated on the difficulty axis ($p=0.23$). Poor-discriminating items $r_{pb}<.10$ appear in six of the seven question types, with the greatest concentration in true/false (24 items) and MCQ radio (17 items). Only select errors has no poor items.

\subsection{Image presence}

The results report a consistently higher raw accuracy on image-present items for eleven of the twelve models, with differences ranging from $+6.0$ pp (Mistral Large 3) to $+18.5$ pp (Claude Sonnet 4.6). The only exception is Pixtral Large scoring -9.2 percentage points lower on image-present items. However, the two subsets, one containing 435 image-absent and the other 436 image-present items, are not matched on type and language. The observed differences may reflect subset composition as well.

\begin{table}[hbt!]
\centering
\caption{Logistic regression predicting item-level accuracy
         ($n = 10{,}452$ observations; reference: MCQ radio,
         no image, English, Gemini 3.1 Flash. Source: Authors' computation}
\label{tab:logistic_regression}
\begin{tabular}{l r r r}
\toprule
\textbf{Predictor} & \textbf{OR} & \textbf{95 percent CI} & \textbf{\textit{p}} \\
\midrule
\multicolumn{4}{l}{\textit{Question type (ref.\ MCQ radio)}} \\
\quad MCQ check         & 1.655 & [1.35,\ 2.04] & $<$.001 \\
\quad True/false        & 0.939 & [0.77,\ 1.15] & .546    \\
\quad Positioning       & 0.727 & [0.60,\ 0.88] & $<$.001 \\
\quad Completion (cl.)  & 0.256 & [0.21,\ 0.31] & $<$.001 \\
\quad Completion (op.)  & 0.432 & [0.35,\ 0.53] & $<$.001 \\
\quad Select errors     & 0.045 & [0.04,\ 0.06] & $<$.001 \\
\midrule
\multicolumn{4}{l}{\textit{Item features}} \\
\quad \textbf{Image present} & \textbf{0.765} & \textbf{[0.67,\ 0.88]} & $\mathbf{<}$\textbf{.001} \\
\quad Language: Italian & 0.459 & [0.38,\ 0.56] & $<$.001 \\
\midrule
\multicolumn{4}{l}{\textit{Model fixed effects (ref.\ Gemini~3.1~Flash~Lite~Preview)}} \\
\quad GPT-5.5                & 3.374 & [2.45,\ 4.64] & $<$.001 \\
\quad Gemini 3.5 Flash       & 1.562 & [1.19,\ 2.06] & .002    \\
\quad Claude Sonnet 4.6      & 1.308 & [1.00,\ 1.71] & .049    \\
\quad Claude Opus 4.6        & 1.133 & [0.87,\ 1.47] & .350    \\
\quad Qwen3-VL-235B          & 1.123 & [0.86,\ 1.46] & .386    \\
\quad Mistral Large 3 675B   & 0.712 & [0.56,\ 0.91] & .007    \\
\quad GPT-5.4 Mini           & 0.653 & [0.51,\ 0.83] & $<$.001 \\
\quad Claude Haiku 4.5       & 0.526 & [0.41,\ 0.67] & $<$.001 \\
\quad GPT-5.4 Nano           & 0.241 & [0.19,\ 0.30] & $<$.001 \\
\quad Pixtral Large          & 0.096 & [0.08,\ 0.12] & $<$.001 \\
\bottomrule
\end{tabular}
\end{table}

Table~\ref{tab:logistic_regression} reports the logistic regression estimating the independent contribution of image presence, controlling for model, question type, and language. The image presence coefficient is negative. The image-presence coefficient is negative, with an odds ratio (OR) of 0.765, a 95 percent confidence interval ranging from 0.67 to 0.88, and a low p-value (p < .001). This indicates that, after controlling for all other variables, a question with an associated image has 76.3 percent of the odds of being answered correctly relative to a text-only question.

\subsection{Question formats}

\begin{figure}
    \centering
    \includegraphics[width=1\linewidth]{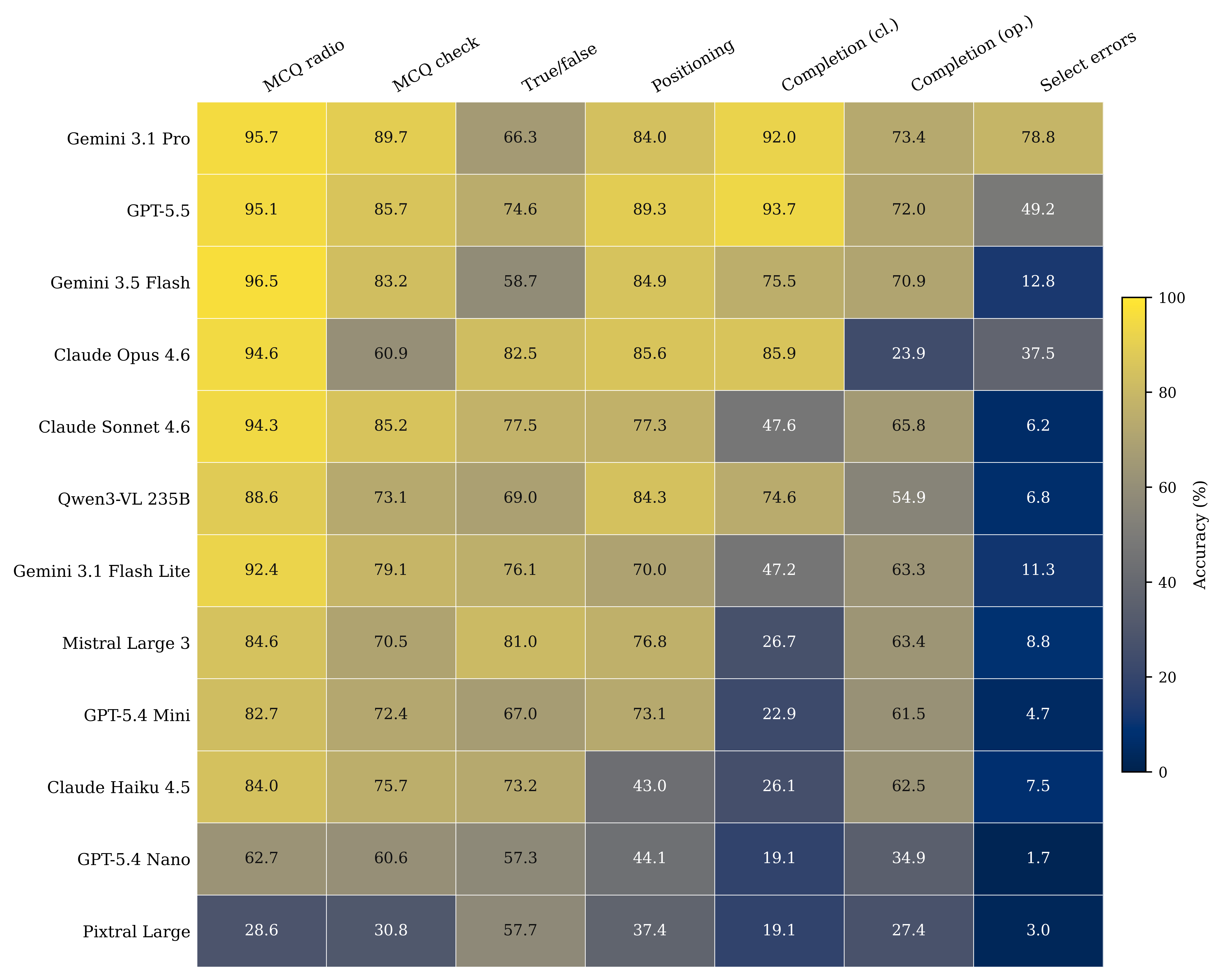}
    \caption{Accuracy of each model across the seven question formats. Models are ordered by macro-average score.}
    \label{fig:question–heatmap}
\end{figure}

Figure~\ref{fig:question–heatmap} reports accuracy per model across the seven question formats proving that no model ranks consistently across formats. On positioning, GPT-5.5 leads with 89.3 percent accuracy. Gemini 3.1 Pro Preview leads on select errors (78.8 percent) while GPT-5.4 Nano only scores 1.7 percent. Gemini 3.5 Flash leads on MCQ radio (96.5 percent) but scores 58.7 percent on true/false. The spread between the best and worst format performance exceeds 50 percentage points for 10 models, with only Gemini 3.1 Pro Preview (29.4 pp) and GPT 5.5 (45.9 pp) showing narrower spreads.

The logistic regression (Table~\ref{tab:logistic_regression}) show that most of the non-reference formats yeld significant coefficients. Most non-reference formats yield significantly lower odds than MCQ radio: positioning (OR = 0.727, p < .001), completion open (OR = 0.432, p < .001), completion closed (OR = 0.256, p < .001), and select errors (OR = 0.045, p < .001). MCQ check is the only format with higher odds (OR = 1.655, p < .001). True/false does not differ significantly from MCQ radio (OR = 0.939, p = .546).

\section{Discussion}

This section discusses EduArt empirical findings in relation to its three research questions. The first concerns whether an educational-level benchmark serves as a valid measurement instrument for multimodal LLMs. The second examines whether question format is an independent source of variance. The third investigates whether requiring a model to justify its answer affects accuracy. Each is addressed in
turn, relating the findings to the gaps in art-historical and multimodal benchmarking identified earlier.

Across the benchmark, the sources of difficulty are not uniform. Table~\ref{tab:model_ranking} clearly highlights this pattern, with MCQ showing a large cluster of high-accuracy levels reached by six models, ranging from 92 percent to 97 percent, with overlapping CIs. This situation mirrors the saturation observed in general-purpose benchmarks such as MMLU, where top models converge toward ceiling performance, eliminating the diagnostic signal of this question format. Moreover, MCQ radio is the type with the highest mean discrimination ($r_{pb} = 0.608$), but its discriminating items separate weak models from capable ones rather than distinguishing among the strongest. EduArt also exhibits the same phenomenon observed by \citep{polo_tinybenchmarks_2024}, who showed that most items carry no diagnostic signal once strong models converge near the ceiling. This is also supported by the existence of 37 MCQ radio items that are correctly answered by all 12 models and therefore they show no variance. 

As a result, these conditions together motivate the inclusion of other formats beyond multiple-choice. There is a need for non-MCQ formats to effectively diagnose specific capabilities in frontier models in this domain. Free-format answer generation, open-ended description, or comparative analysis tasks are natural candidates for future evaluation designs that can preserve discriminative power as model capabilities continue to improve. On the other hand, these formats require much more effort on the evaluation task. Overall, Item difficulty is well-distributed across the full range (mean $p = 0.696$, SD $= 0.232$), confirming that the benchmark is not trivial or systematically impossible. Discrimination is strong for the large majority of items ($\bar{r}_{pb} = 0.514$, with 82.3 percent classified as good discriminators). These properties are in line with prior item-level evaluations, and they are deemed as indicative of well-functioning instruments \citep{polo_tinybenchmarks_2024,zhou_lost_2026}.

An observable pattern across the formats is that the models do not rank consistently, and the spread between a model's best and worst formats exceeds 50 percentage points for 10 of the 12 models. Models scoring above 94 percent on MCQ radio can fall  lower on formats that require producing rather than recognizing art-historical content (e.g., Claude Sonnet 4.6 at 6.2 percent on select errors). This indicates that possessing knowledge and being able to deploy it through different task structures are distinct capabilities.

The logistic regression further isolates the formats. The coefficients indicate that the MCQ check is the only format with a higher likelihood of receiving a correct response than the MCQ radio. This is likely attributable to the partial-credit nature of F1 scoring rather than to a genuine simplicity. In fact, this format may inflate the score due to partial matches, compared to the stricter exact-match format used in the MCQ radio format. The remaining formats yield lower odds, and their overall decreasing accuracies reflect an increase in art-historical knowledge rather than recognition from a closed set (Fig. \ref{fig:question–heatmap}), proving that current multimodal LLMs are still struggling on this task. This gap connects to a broader pattern seen in the
visual reasoning literature. \citep{zellers_recognition_2019} distinguished recognition from cognition, arguing that genuine comprehension requires inferences beyond recognition of surface content. \citep{zhang_mathverse_2025} empirically showed that multimodal LLMs often rely on language-based reasoning that masquerades as visual reasoning, with performance degrading only when the shortcuts are explicitly blocked. In the EduArt case, the non-MCQ formats act as shortcut-blocking mechanisms in the art-historical domain, removing the option set and forcing the model to generate, correctly place, or evaluate content directly. 

An exception to the informative nature of non-MCQ questions is the true/false typology. The format presents a probability score similar to that of the MCQ Radio (Fig. \ref{fig:difficulty}) and the lowest mean discrimination from the CTT analysis (Tab. \ref{tab:item_discrimination}). This is due to statement-level scoring, which allows the models to accumulate partial credit instead of proving genuine art-historical ability. Therefore, the format contributes little diagnostic value beyond what MCQ radio already captures.

Taken together, these results have direct implications for benchmark design in the cultural heritage domain. A model's MCQ performance does not predict its performance on completion or positioning, which means that single-format evaluations systematically overestimate the range of tasks a model can reliably perform. For applications in art history, where tasks frequently require producing and interpreting art-historical content rather than selecting from options, MCQ-based benchmarks offer a misleadingly optimistic picture of the model's capabilities.

Regarding image presence, raw accuracy is higher on image-present items, which is counterintuitive for a benchmark grounded in visual materials, as adding visual information is expected to increase the difficulty and the joint use of both modalities. Analyzing the data using the logistic regressions (Tab. \ref{tab:logistic_regression}), after controlling for question type, language, and model, image presence is associated with lower odds of a correct response. Therefore, images make items harder, as expected for a domain where visual interpretation is a genuine source of difficulty, and consistent with the view that language-based knowledge rather than visual processing drives most of the accuracy observed on EduArt \citep{zhang_mathverse_2025}. 

\begin{figure}
    \centering
    \includegraphics[width=1\linewidth]{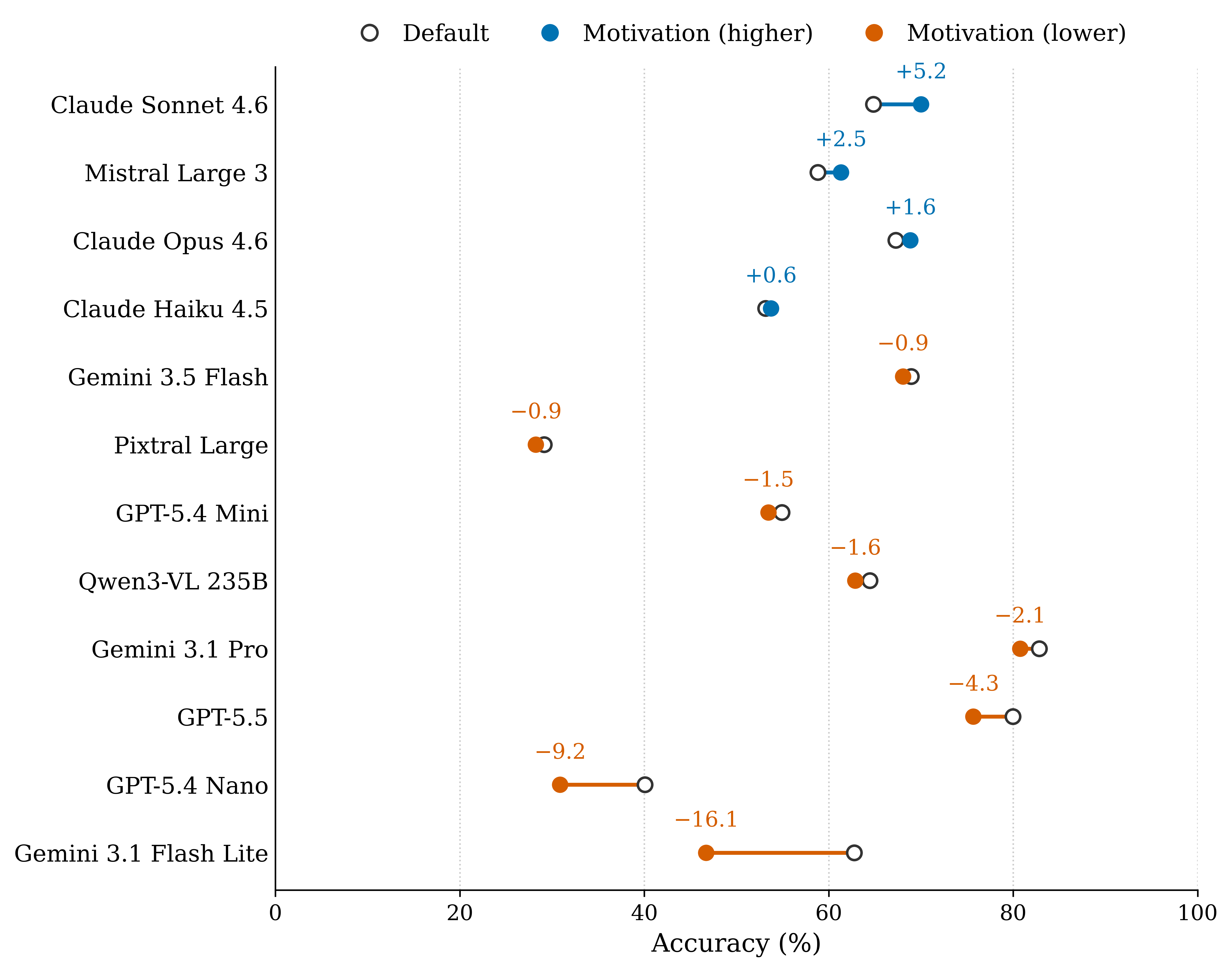}
    \caption{Change in macro-average accuracy from the default condition to the
motivation condition for each model. Open circles show default accuracy, filled
circles show motivation accuracy, with blue marks showing an increase, red a decrease.}
    \label{fig:motivation}
\end{figure}

The motivation condition produces a clear family-level pattern, suggesting that model families systematically differ in accuracy when required to articulate a justification alongside their answer. Figure~\ref{fig:motivation} shows that all three Claude models improve, while both Gemini models and all three GPT models degrade. The largest degradation is in Gemini 3.1 Flash Lite Preview (−12.8 pp), followed by GPT-5.4 Nano (−6.7 pp), while the Claude gains are more modest, led by Claude Sonnet 4.6 (+2.6 pp). Mistral Large 3 also improves slightly (+0.4 pp), whereas Qwen3-VL-235B (−1.8 pp) and Pixtral Large (−1.9 pp) show small declines. This could mean that requiring a model to justify its choice alongside the answer interferes with the answer-selection process, particularly in formats that demand precise placement or text generation. Distinguishing this from other possible mechanisms, however, would require access to model internals beyond what a black-box evaluation can provide, which is outside this project's scope.

\begin{table}[hbt!]
\centering
\caption{Effect of the motivation condition by question type. Mean paired
difference (motivation $-$ default) in percentage points, averaged across the
twelve models. Positive values indicate higher accuracy under the
motivation condition.
Source: Authors' computation.}
\label{tab:motivation_by_type}
\begin{tabular}{l r r}
\toprule
\textbf{Question Type} & \textbf{Mean $\Delta$ (pp)} & \textbf{Range (pp)} \\
\midrule
Completion (closed) & $+2.0$ & $[-35.0,\ +40.9]$ \\
Select errors       & $-0.7$ & $[-13.1,\ +16.1]$ \\
MCQ radio           & $-0.9$ & $[-4.0,\ +1.1]$  \\
Completion (open)   & $-2.7$ & $[-16.6,\ +16.0]$ \\
MCQ check           & $-4.1$ & $[-33.3,\ +5.1]$ \\
True/false          & $-4.6$ & $[-14.1,\ +0.8]$ \\
Positioning         & $-4.6$ & $[-26.7,\ +12.9]$ \\
\bottomrule
\end{tabular}
\end{table}

The type-level breakdown, in Tab. \ref{tab:motivation_by_type}, shows that the motivation effect is not uniform across formats. The largest gains concentrate in the “completion closed” tasks, which require precise recall of the correct words to fill blanks in text. Conversely, the largest degradations occur in formats requiring structured spatial responses, such as positioning, where the justification requirement appears to act as a distractor rather than a support for the answer-generation process. The family-level split is the clearest signal: the Claude models are the only ones to benefit from elicited justification, while every Gemini and GPT model declines, suggesting that the interaction between justification and accuracy depends on how each family is trained to handle simultaneous reasoning and structured output.

\section{Conclusion}

EduArt demonstrates that educational-level art history questions, drawn from human-designed assessment materials and spanning different question formats, constitute a valid and non-trivial benchmark for evaluating multimodal LLMs. Its manually authored dataset sets it apart from prior art-focused benchmarks built on synthetically generated content, and the CTT item analysis confirms strong psychometric properties. An adequate difficulty spread and a mean discrimination of 0.514 with 82.3 percent of items classified as good discriminators. By reporting these item-level statistics, absent from earlier benchmarks, EduArt enables attributing model failure to genuine difficulty rather than to shortcuts exploited inherited by the question construction. The results show that even frontier models are far from saturating the benchmark as a whole, with macro-average scores ranging from 29.1 percent to 82.8 percent and near-floor performance on select error typologies for most of the evaluated models. The central finding across the three research questions is that art-historical Knowledge and the ability to demonstrate it are note expressed in the same way. Models that exceed 94 percent on multiple-choice recognition can fall below 25 percent on open completion or below 10 percent on error identification in art-historical text. Requiring a model to justify its answer as a structured-output manipulation changes accuracy predominantly in negative, family-dependent ways. These dissociations are diagnostic signals indicating what the current multimodal models can and cannot do in a specialized domain.

Currently, generative multimodal models are used to support specific tasks, such as coding, knowledge extraction, and more. If generative multimodal models are to support art-historical scholarship in tasks such as catalogation, attribution, iconographic description, or teaching, knowing their performance in multiple-choice questions is not enough. What matters is knowing precisely which tasks they handle reliably
and which they do not, because the tasks that arise in scholarly work rarely take the form of selecting an answer from a fixed set. Establishing this fine-grained map of capability is therefore a precondition for responsible adoption: it allows the field to direct these tools toward the uses they genuinely support while
guarding against the ones they do not.

\paragraph{Acknowledgments}

During the preparation of this manuscript, the authors used the following AI tools for the purposes described.
In the research pipeline (described in Methods). Google Gemini 3 Flash (gemini-3-flash, accessed March 2026) and Anthropic Claude Sonnet 4.5 (claude-sonnet-4-5-20250929, accessed March 2026) were used to extract structured question-answer records from screenshots and HTML of the MyZanichelli platform. The full extraction prompt is available at \url{https://anonymous.4open.science/r/EduArt-educational-level-benchmark/prompts/answer_question.txt}. All extractions were independently produced by both models, compared programmatically, and reviewed manually before inclusion in the dataset.
In manuscript preparation. OpenAI ChatGPT (GPT-5) and Anthropic Claude (Opus 4.x) were used to support drafting of the Results section and to refine the prose of all sections. The authors reviewed and edited all AI-assisted passages and take full responsibility for the accuracy of claims, interpretations, citations, and the final wording of the manuscript.  The authors also thank Dr. Lisandra S. Costiner, Utrecht University, for suggesting the use of the AP Art History examinations as the English-language source for the benchmark.

\paragraph{Funding Statement}

This research was supported by Villa I Tatti — The Harvard University Center for Italian Renaissance Studies, which provided API access to Anthropic, OpenAI, Qwen, and Mistral models via the Harvard Bedrock account. Access to Google Gemini models was funded through credits granted to the authors under the Google Academic Research Credits Programme. The funders had no role in study design, data collection and analysis, decision to publish, or preparation of the manuscript.

\paragraph{Data Availability Statement}
The benchmark items themselves are educational materials protected by copyright and cannot be redistributed. The following resources are openly available in the project's GitHub repository at \url{https://anonymous.4open.science/r/EduArt-educational-level-benchmark}: (i) the full extraction and evaluation pipeline; (ii) the system and user prompts used for question extraction and model evaluation; (iii) item-level model outputs and graded scores for all 12 models under both experimental conditions; (iv) the psychometric analysis code (CTT and logistic regression) and figure-generation notebooks. Access to the source materials requires a MyZanichelli account (Zanichelli, Italy) and download of released AP Art History examinations from the College Board website.

%\endnote in some journals will behave like \footnote; and \printendnotes will not output anything. 
%\printendnotes

\bibliographystyle{unsrtnat}
\bibliography{references}  %%% Uncomment this line and comment out the ``thebibliography'' section below to use the external .bib file (using bibtex) .

@misc{hendrycks_measuring_2021,
	title = {Measuring Massive Multitask Language Understanding},
	url = {http://arxiv.org/abs/2009.03300},
	doi = {10.48550/arXiv.2009.03300},
	number = {{arXiv}:2009.03300},
	publisher = {{arXiv}},
	author = {Hendrycks, Dan and Burns, Collin and Basart, Steven and Zou, Andy and Mazeika, Mantas and Song, Dawn and Steinhardt, Jacob},
	urldate = {2026-03-23},
	date = {2021-01-12},
	eprinttype = {arxiv},
	eprint = {2009.03300 [cs]},
}

@article{wang_mmlu-pro_2024,
	title = {{MMLU}-Pro: A More Robust and Challenging Multi-Task Language Understanding Benchmark},
	volume = {37},
	url = {https://proceedings.neurips.cc/paper_files/paper/2024/hash/ad236edc564f3e3156e1b2feafb99a24-Abstract-Datasets_and_Benchmarks_Track.html},
	doi = {10.52202/079017-3018},
	shorttitle = {{MMLU}-Pro},
	pages = {95266--95290},
	journaltitle = {Advances in Neural Information Processing Systems},
	author = {Wang, Yubo and Ma, Xueguang and Zhang, Ge and Ni, Yuansheng and Chandra, Abhranil and Guo, Shiguang and Ren, Weiming and Arulraj, Aaran and He, Xuan and Jiang, Ziyan and Li, Tianle and Ku, Max and Wang, Kai and Zhuang, Alex and Fan, Rongqi and Yue, Xiang and Chen, Wenhu},
	urldate = {2026-03-23},
	date = {2024-12-16},
	langid = {english},
}

@inproceedings{rein_gpqa_2024,
	title = {{GPQA}: A Graduate-Level Google-Proof Q\&A Benchmark},
	url = {https://openreview.net/forum?id=Ti67584b98&utm_campaign=The%20Batch&utm_source=hs_email&utm_medium=email},
	shorttitle = {{GPQA}},
	eventtitle = {First Conference on Language Modeling},
	author = {Rein, David and Hou, Betty Li and Stickland, Asa Cooper and Petty, Jackson and Pang, Richard Yuanzhe and Dirani, Julien and Michael, Julian and Bowman, Samuel R.},
	urldate = {2026-03-23},
	date = {2024-08-26},
	langid = {english},
}

@article{guha_legalbench_2023,
	title = {{LegalBench}: A Collaboratively Built Benchmark for Measuring Legal Reasoning in Large Language Models},
	volume = {36},
	url = {https://proceedings.neurips.cc/paper_files/paper/2023/hash/89e44582fd28ddfea1ea4dcb0ebbf4b0-Abstract-Datasets_and_Benchmarks.html},
	shorttitle = {{LegalBench}},
	pages = {44123--44279},
	journaltitle = {Advances in Neural Information Processing Systems},
	author = {Guha, Neel and Nyarko, Julian and Ho, Daniel and Ré, Christopher and Chilton, Adam and K, Aditya and Chohlas-Wood, Alex and Peters, Austin and Waldon, Brandon and Rockmore, Daniel and Zambrano, Diego and Talisman, Dmitry and Hoque, Enam and Surani, Faiz and Fagan, Frank and Sarfaty, Galit and Dickinson, Gregory and Porat, Haggai and Hegland, Jason and Wu, Jessica and Nudell, Joe and Niklaus, Joel and Nay, John and Choi, Jonathan and Tobia, Kevin and Hagan, Margaret and Ma, Megan and Livermore, Michael and Rasumov-Rahe, Nikon and Holzenberger, Nils and Kolt, Noam and Henderson, Peter and Rehaag, Sean and Goel, Sharad and Gao, Shang and Williams, Spencer and Gandhi, Sunny and Zur, Tom and Iyer, Varun and Li, Zehua},
	urldate = {2026-03-23},
	date = {2023-12-15},
	langid = {english},
}

@article{lu_learn_2022,
	title = {Learn to Explain: Multimodal Reasoning via Thought Chains for Science Question Answering},
	volume = {35},
	url = {https://proceedings.neurips.cc/paper_files/paper/2022/hash/11332b6b6cf4485b84afadb1352d3a9a-Abstract-Conference.html},
	shorttitle = {Learn to Explain},
	pages = {2507--2521},
	journaltitle = {Advances in Neural Information Processing Systems},
	author = {Lu, Pan and Mishra, Swaroop and Xia, Tanglin and Qiu, Liang and Chang, Kai-Wei and Zhu, Song-Chun and Tafjord, Oyvind and Clark, Peter and Kalyan, Ashwin},
	urldate = {2026-03-23},
	date = {2022-12-06},
	langid = {english},
}

@article{chang_survey_2024,
	title = {A Survey on Evaluation of Large Language Models},
	volume = {15},
	issn = {2157-6904},
	url = {https://dl.acm.org/doi/10.1145/3641289},
	doi = {10.1145/3641289},
	pages = {39:1--39:45},
	number = {3},
	journaltitle = {{ACM} Trans. Intell. Syst. Technol.},
	author = {Chang, Yupeng and Wang, Xu and Wang, Jindong and Wu, Yuan and Yang, Linyi and Zhu, Kaijie and Chen, Hao and Yi, Xiaoyuan and Wang, Cunxiang and Wang, Yidong and Ye, Wei and Zhang, Yue and Chang, Yi and Yu, Philip S. and Yang, Qiang and Xie, Xing},
	urldate = {2026-03-23},
	date = {2024-03-29},
}

@inproceedings{yue_mmmu_2024,
	title = {{MMMU}: A Massive Multi-discipline Multimodal Understanding and Reasoning Benchmark for Expert {AGI}},
	url = {https://openaccess.thecvf.com/content/CVPR2024/html/Yue_MMMU_A_Massive_Multi-discipline_Multimodal_Understanding_and_Reasoning_Benchmark_for_CVPR_2024_paper.html},
	shorttitle = {{MMMU}},
	eventtitle = {Proceedings of the {IEEE}/{CVF} Conference on Computer Vision and Pattern Recognition},
	pages = {9556--9567},
	author = {Yue, Xiang and Ni, Yuansheng and Zhang, Kai and Zheng, Tianyu and Liu, Ruoqi and Zhang, Ge and Stevens, Samuel and Jiang, Dongfu and Ren, Weiming and Sun, Yuxuan and Wei, Cong and Yu, Botao and Yuan, Ruibin and Sun, Renliang and Yin, Ming and Zheng, Boyuan and Yang, Zhenzhu and Liu, Yibo and Huang, Wenhao and Sun, Huan and Su, Yu and Chen, Wenhu},
	urldate = {2026-03-23},
	date = {2024},
	langid = {english},
}

@inproceedings{alfarano_vqart-bench_2025,
	title = {{VQArt}-Bench: A Semantically Rich {VQA} Benchmark for Art and Cultural Heritage},
	issn = {2473-9944},
	url = {https://ieeexplore.ieee.org/abstract/document/11375590},
	doi = {10.1109/ICCVW69036.2025.00761},
	shorttitle = {{VQArt}-Bench},
	eventtitle = {2025 {IEEE}/{CVF} International Conference on Computer Vision Workshops ({ICCVW})},
	pages = {406--416},
	booktitle = {2025 {IEEE}/{CVF} International Conference on Computer Vision Workshops ({ICCVW})},
	author = {Alfarano, Andrea and Venturoli, Lorenzo and del Castillo, Darío Negueruela},
	urldate = {2026-03-23},
	date = {2025-10},
	note = {{ISSN}: 2473-9944},
}

@misc{openai_gpt-4_2024,
	title = {{GPT}-4 Technical Report},
	url = {http://arxiv.org/abs/2303.08774},
	doi = {10.48550/arXiv.2303.08774},
	number = {{arXiv}:2303.08774},
	publisher = {{arXiv}},
	author = {{OpenAI} and Achiam, Josh and Adler, Steven and Agarwal, Sandhini and Ahmad, Lama and Akkaya, Ilge and Aleman, Florencia Leoni and Almeida, Diogo and Altenschmidt, Janko and Altman, Sam and Anadkat, Shyamal and Avila, Red and Babuschkin, Igor and Balaji, Suchir and Balcom, Valerie and Baltescu, Paul and Bao, Haiming and Bavarian, Mohammad and Belgum, Jeff and Bello, Irwan and Berdine, Jake and Bernadett-Shapiro, Gabriel and Berner, Christopher and Bogdonoff, Lenny and Boiko, Oleg and Boyd, Madelaine and Brakman, Anna-Luisa and Brockman, Greg and Brooks, Tim and Brundage, Miles and Button, Kevin and Cai, Trevor and Campbell, Rosie and Cann, Andrew and Carey, Brittany and Carlson, Chelsea and Carmichael, Rory and Chan, Brooke and Chang, Che and Chantzis, Fotis and Chen, Derek and Chen, Sully and Chen, Ruby and Chen, Jason and Chen, Mark and Chess, Ben and Cho, Chester and Chu, Casey and Chung, Hyung Won and Cummings, Dave and Currier, Jeremiah and Dai, Yunxing and Decareaux, Cory and Degry, Thomas and Deutsch, Noah and Deville, Damien and Dhar, Arka and Dohan, David and Dowling, Steve and Dunning, Sheila and Ecoffet, Adrien and Eleti, Atty and Eloundou, Tyna and Farhi, David and Fedus, Liam and Felix, Niko and Fishman, Simón Posada and Forte, Juston and Fulford, Isabella and Gao, Leo and Georges, Elie and Gibson, Christian and Goel, Vik and Gogineni, Tarun and Goh, Gabriel and Gontijo-Lopes, Rapha and Gordon, Jonathan and Grafstein, Morgan and Gray, Scott and Greene, Ryan and Gross, Joshua and Gu, Shixiang Shane and Guo, Yufei and Hallacy, Chris and Han, Jesse and Harris, Jeff and He, Yuchen and Heaton, Mike and Heidecke, Johannes and Hesse, Chris and Hickey, Alan and Hickey, Wade and Hoeschele, Peter and Houghton, Brandon and Hsu, Kenny and Hu, Shengli and Hu, Xin and Huizinga, Joost and Jain, Shantanu and Jain, Shawn and Jang, Joanne and Jiang, Angela and Jiang, Roger and Jin, Haozhun and Jin, Denny and Jomoto, Shino and Jonn, Billie and Jun, Heewoo and Kaftan, Tomer and Kaiser, Łukasz and Kamali, Ali and Kanitscheider, Ingmar and Keskar, Nitish Shirish and Khan, Tabarak and Kilpatrick, Logan and Kim, Jong Wook and Kim, Christina and Kim, Yongjik and Kirchner, Jan Hendrik and Kiros, Jamie and Knight, Matt and Kokotajlo, Daniel and Kondraciuk, Łukasz and Kondrich, Andrew and Konstantinidis, Aris and Kosic, Kyle and Krueger, Gretchen and Kuo, Vishal and Lampe, Michael and Lan, Ikai and Lee, Teddy and Leike, Jan and Leung, Jade and Levy, Daniel and Li, Chak Ming and Lim, Rachel and Lin, Molly and Lin, Stephanie and Litwin, Mateusz and Lopez, Theresa and Lowe, Ryan and Lue, Patricia and Makanju, Anna and Malfacini, Kim and Manning, Sam and Markov, Todor and Markovski, Yaniv and Martin, Bianca and Mayer, Katie and Mayne, Andrew and {McGrew}, Bob and {McKinney}, Scott Mayer and {McLeavey}, Christine and {McMillan}, Paul and {McNeil}, Jake and Medina, David and Mehta, Aalok and Menick, Jacob and Metz, Luke and Mishchenko, Andrey and Mishkin, Pamela and Monaco, Vinnie and Morikawa, Evan and Mossing, Daniel and Mu, Tong and Murati, Mira and Murk, Oleg and Mély, David and Nair, Ashvin and Nakano, Reiichiro and Nayak, Rajeev and Neelakantan, Arvind and Ngo, Richard and Noh, Hyeonwoo and Ouyang, Long and O'Keefe, Cullen and Pachocki, Jakub and Paino, Alex and Palermo, Joe and Pantuliano, Ashley and Parascandolo, Giambattista and Parish, Joel and Parparita, Emy and Passos, Alex and Pavlov, Mikhail and Peng, Andrew and Perelman, Adam and Peres, Filipe de Avila Belbute and Petrov, Michael and Pinto, Henrique Ponde de Oliveira and Michael and Pokorny and Pokrass, Michelle and Pong, Vitchyr H. and Powell, Tolly and Power, Alethea and Power, Boris and Proehl, Elizabeth and Puri, Raul and Radford, Alec and Rae, Jack and Ramesh, Aditya and Raymond, Cameron and Real, Francis and Rimbach, Kendra and Ross, Carl and Rotsted, Bob and Roussez, Henri and Ryder, Nick and Saltarelli, Mario and Sanders, Ted and Santurkar, Shibani and Sastry, Girish and Schmidt, Heather and Schnurr, David and Schulman, John and Selsam, Daniel and Sheppard, Kyla and Sherbakov, Toki and Shieh, Jessica and Shoker, Sarah and Shyam, Pranav and Sidor, Szymon and Sigler, Eric and Simens, Maddie and Sitkin, Jordan and Slama, Katarina and Sohl, Ian and Sokolowsky, Benjamin and Song, Yang and Staudacher, Natalie and Such, Felipe Petroski and Summers, Natalie and Sutskever, Ilya and Tang, Jie and Tezak, Nikolas and Thompson, Madeleine B. and Tillet, Phil and Tootoonchian, Amin and Tseng, Elizabeth and Tuggle, Preston and Turley, Nick and Tworek, Jerry and Uribe, Juan Felipe Cerón and Vallone, Andrea and Vijayvergiya, Arun and Voss, Chelsea and Wainwright, Carroll and Wang, Justin Jay and Wang, Alvin and Wang, Ben and Ward, Jonathan and Wei, Jason and Weinmann, C. J. and Welihinda, Akila and Welinder, Peter and Weng, Jiayi and Weng, Lilian and Wiethoff, Matt and Willner, Dave and Winter, Clemens and Wolrich, Samuel and Wong, Hannah and Workman, Lauren and Wu, Sherwin and Wu, Jeff and Wu, Michael and Xiao, Kai and Xu, Tao and Yoo, Sarah and Yu, Kevin and Yuan, Qiming and Zaremba, Wojciech and Zellers, Rowan and Zhang, Chong and Zhang, Marvin and Zhao, Shengjia and Zheng, Tianhao and Zhuang, Juntang and Zhuk, William and Zoph, Barret},
	urldate = {2026-03-30},
	date = {2024-03-04},
	eprinttype = {arxiv},
	eprint = {2303.08774 [cs]},
}

@misc{team_gemini_2025,
	title = {Gemini: A Family of Highly Capable Multimodal Models},
	url = {http://arxiv.org/abs/2312.11805},
	doi = {10.48550/arXiv.2312.11805},
	shorttitle = {Gemini},
	number = {{arXiv}:2312.11805},
	publisher = {{arXiv}},
	author = {Gemini and Anil, Rohan and Borgeaud, Sebastian and Alayrac, Jean-Baptiste and Yu, Jiahui and Soricut, Radu and Schalkwyk, Johan and Dai, Andrew M. and Hauth, Anja and Millican, Katie and Silver, David and Johnson, Melvin and Antonoglou, Ioannis and Schrittwieser, Julian and Glaese, Amelia and Chen, Jilin and Pitler, Emily and Lillicrap, Timothy and Lazaridou, Angeliki and Firat, Orhan and Molloy, James and Isard, Michael and Barham, Paul R. and Hennigan, Tom and Lee, Benjamin and Viola, Fabio and Reynolds, Malcolm and Xu, Yuanzhong and Doherty, Ryan and Collins, Eli and Meyer, Clemens and Rutherford, Eliza and Moreira, Erica and Ayoub, Kareem and Goel, Megha and Krawczyk, Jack and Du, Cosmo and Chi, Ed and Cheng, Heng-Tze and Ni, Eric and Shah, Purvi and Kane, Patrick and Chan, Betty and Faruqui, Manaal and Severyn, Aliaksei and Lin, Hanzhao and Li, {YaGuang} and Cheng, Yong and Ittycheriah, Abe and Mahdieh, Mahdis and Chen, Mia and Sun, Pei and Tran, Dustin and Bagri, Sumit and Lakshminarayanan, Balaji and Liu, Jeremiah and Orban, Andras and Güra, Fabian and Zhou, Hao and Song, Xinying and Boffy, Aurelien and Ganapathy, Harish and Zheng, Steven and Choe, {HyunJeong} and Weisz, Ágoston and Zhu, Tao and Lu, Yifeng and Gopal, Siddharth and Kahn, Jarrod and Kula, Maciej and Pitman, Jeff and Shah, Rushin and Taropa, Emanuel and Merey, Majd Al and Baeuml, Martin and Chen, Zhifeng and Shafey, Laurent El and Zhang, Yujing and Sercinoglu, Olcan and Tucker, George and Piqueras, Enrique and Krikun, Maxim and Barr, Iain and Savinov, Nikolay and Danihelka, Ivo and Roelofs, Becca and White, Anaïs and Andreassen, Anders and Glehn, Tamara von and Yagati, Lakshman and Kazemi, Mehran and Gonzalez, Lucas and Khalman, Misha and Sygnowski, Jakub and Frechette, Alexandre and Smith, Charlotte and Culp, Laura and Proleev, Lev and Luan, Yi and Chen, Xi and Lottes, James and Schucher, Nathan and Lebron, Federico and Rrustemi, Alban and Clay, Natalie and Crone, Phil and Kocisky, Tomas and Zhao, Jeffrey and Perz, Bartek and Yu, Dian and Howard, Heidi and Bloniarz, Adam and Rae, Jack W. and Lu, Han and Sifre, Laurent and Maggioni, Marcello and Alcober, Fred and Garrette, Dan and Barnes, Megan and Thakoor, Shantanu and Austin, Jacob and Barth-Maron, Gabriel and Wong, William and Joshi, Rishabh and Chaabouni, Rahma and Fatiha, Deeni and Ahuja, Arun and Tomar, Gaurav Singh and Senter, Evan and Chadwick, Martin and Kornakov, Ilya and Attaluri, Nithya and Iturrate, Iñaki and Liu, Ruibo and Li, Yunxuan and Cogan, Sarah and Chen, Jeremy and Jia, Chao and Gu, Chenjie and Zhang, Qiao and Grimstad, Jordan and Hartman, Ale Jakse and Garcia, Xavier and Pillai, Thanumalayan Sankaranarayana and Devlin, Jacob and Laskin, Michael and Casas, Diego de Las and Valter, Dasha and Tao, Connie and Blanco, Lorenzo and Badia, Adrià Puigdomènech and Reitter, David and Chen, Mianna and Brennan, Jenny and Rivera, Clara and Brin, Sergey and Iqbal, Shariq and Surita, Gabriela and Labanowski, Jane and Rao, Abhi and Winkler, Stephanie and Parisotto, Emilio and Gu, Yiming and Olszewska, Kate and Addanki, Ravi and Miech, Antoine and Louis, Annie and Teplyashin, Denis and Brown, Geoff and Catt, Elliot and Balaguer, Jan and Xiang, Jackie and Wang, Pidong and Ashwood, Zoe and Briukhov, Anton and Webson, Albert and Ganapathy, Sanjay and Sanghavi, Smit and Kannan, Ajay and Chang, Ming-Wei and Stjerngren, Axel and Djolonga, Josip and Sun, Yuting and Bapna, Ankur and Aitchison, Matthew and Pejman, Pedram and Michalewski, Henryk and Yu, Tianhe and Wang, Cindy and Love, Juliette and Ahn, Junwhan and Bloxwich, Dawn and Han, Kehang and Humphreys, Peter and Sellam, Thibault and Bradbury, James and Godbole, Varun and Samangooei, Sina and Damoc, Bogdan and Kaskasoli, Alex and Arnold, Sébastien M. R. and Vasudevan, Vijay and Agrawal, Shubham and Riesa, Jason and Lepikhin, Dmitry and Tanburn, Richard and Srinivasan, Srivatsan and Lim, Hyeontaek and Hodkinson, Sarah and Shyam, Pranav and Ferret, Johan and Hand, Steven and Garg, Ankush and Paine, Tom Le and Li, Jian and Li, Yujia and Giang, Minh and Neitz, Alexander and Abbas, Zaheer and York, Sarah and Reid, Machel and Cole, Elizabeth and Chowdhery, Aakanksha and Das, Dipanjan and Rogozińska, Dominika and Nikolaev, Vitaliy and Sprechmann, Pablo and Nado, Zachary and Zilka, Lukas and Prost, Flavien and He, Luheng and Monteiro, Marianne and Mishra, Gaurav and Welty, Chris and Newlan, Josh and Jia, Dawei and Allamanis, Miltiadis and Hu, Clara Huiyi and Liedekerke, Raoul de and Gilmer, Justin and Saroufim, Carl and Rijhwani, Shruti and Hou, Shaobo and Shrivastava, Disha and Baddepudi, Anirudh and Goldin, Alex and Ozturel, Adnan and Cassirer, Albin and Xu, Yunhan and Sohn, Daniel and Sachan, Devendra and Amplayo, Reinald Kim and Swanson, Craig and Petrova, Dessie and Narayan, Shashi and Guez, Arthur and Brahma, Siddhartha and Landon, Jessica and Patel, Miteyan and Zhao, Ruizhe and Villela, Kevin and Wang, Luyu and Jia, Wenhao and Rahtz, Matthew and Giménez, Mai and Yeung, Legg and Keeling, James and Georgiev, Petko and Mincu, Diana and Wu, Boxi and Haykal, Salem and Saputro, Rachel and Vodrahalli, Kiran and Qin, James and Cankara, Zeynep and Sharma, Abhanshu and Fernando, Nick and Hawkins, Will and Neyshabur, Behnam and Kim, Solomon and Hutter, Adrian and Agrawal, Priyanka and Castro-Ros, Alex and Driessche, George van den and Wang, Tao and Yang, Fan and Chang, Shuo-yiin and Komarek, Paul and {McIlroy}, Ross and Lučić, Mario and Zhang, Guodong and Farhan, Wael and Sharman, Michael and Natsev, Paul and Michel, Paul and Bansal, Yamini and Qiao, Siyuan and Cao, Kris and Shakeri, Siamak and Butterfield, Christina and Chung, Justin and Rubenstein, Paul Kishan and Agrawal, Shivani and Mensch, Arthur and Soparkar, Kedar and Lenc, Karel and Chung, Timothy and Pope, Aedan and Maggiore, Loren and Kay, Jackie and Jhakra, Priya and Wang, Shibo and Maynez, Joshua and Phuong, Mary and Tobin, Taylor and Tacchetti, Andrea and Trebacz, Maja and Robinson, Kevin and Katariya, Yash and Riedel, Sebastian and Bailey, Paige and Xiao, Kefan and Ghelani, Nimesh and Aroyo, Lora and Slone, Ambrose and Houlsby, Neil and Xiong, Xuehan and Yang, Zhen and Gribovskaya, Elena and Adler, Jonas and Wirth, Mateo and Lee, Lisa and Li, Music and Kagohara, Thais and Pavagadhi, Jay and Bridgers, Sophie and Bortsova, Anna and Ghemawat, Sanjay and Ahmed, Zafarali and Liu, Tianqi and Powell, Richard and Bolina, Vijay and Iinuma, Mariko and Zablotskaia, Polina and Besley, James and Chung, Da-Woon and Dozat, Timothy and Comanescu, Ramona and Si, Xiance and Greer, Jeremy and Su, Guolong and Polacek, Martin and Kaufman, Raphaël Lopez and Tokumine, Simon and Hu, Hexiang and Buchatskaya, Elena and Miao, Yingjie and Elhawaty, Mohamed and Siddhant, Aditya and Tomasev, Nenad and Xing, Jinwei and Greer, Christina and Miller, Helen and Ashraf, Shereen and Roy, Aurko and Zhang, Zizhao and Ma, Ada and Filos, Angelos and Besta, Milos and Blevins, Rory and Klimenko, Ted and Yeh, Chih-Kuan and Changpinyo, Soravit and Mu, Jiaqi and Chang, Oscar and Pajarskas, Mantas and Muir, Carrie and Cohen, Vered and Lan, Charline Le and Haridasan, Krishna and Marathe, Amit and Hansen, Steven and Douglas, Sholto and Samuel, Rajkumar and Wang, Mingqiu and Austin, Sophia and Lan, Chang and Jiang, Jiepu and Chiu, Justin and Lorenzo, Jaime Alonso and Sjösund, Lars Lowe and Cevey, Sébastien and Gleicher, Zach and Avrahami, Thi and Boral, Anudhyan and Srinivasan, Hansa and Selo, Vittorio and May, Rhys and Aisopos, Konstantinos and Hussenot, Léonard and Soares, Livio Baldini and Baumli, Kate and Chang, Michael B. and Recasens, Adrià and Caine, Ben and Pritzel, Alexander and Pavetic, Filip and Pardo, Fabio and Gergely, Anita and Frye, Justin and Ramasesh, Vinay and Horgan, Dan and Badola, Kartikeya and Kassner, Nora and Roy, Subhrajit and Dyer, Ethan and Campos, Víctor Campos and Tomala, Alex and Tang, Yunhao and Badawy, Dalia El and White, Elspeth and Mustafa, Basil and Lang, Oran and Jindal, Abhishek and Vikram, Sharad and Gong, Zhitao and Caelles, Sergi and Hemsley, Ross and Thornton, Gregory and Feng, Fangxiaoyu and Stokowiec, Wojciech and Zheng, Ce and Thacker, Phoebe and Ünlü, Çağlar and Zhang, Zhishuai and Saleh, Mohammad and Svensson, James and Bileschi, Max and Patil, Piyush and Anand, Ankesh and Ring, Roman and Tsihlas, Katerina and Vezer, Arpi and Selvi, Marco and Shevlane, Toby and Rodriguez, Mikel and Kwiatkowski, Tom and Daruki, Samira and Rong, Keran and Dafoe, Allan and {FitzGerald}, Nicholas and Gu-Lemberg, Keren and Khan, Mina and Hendricks, Lisa Anne and Pellat, Marie and Feinberg, Vladimir and Cobon-Kerr, James and Sainath, Tara and Rauh, Maribeth and Hashemi, Sayed Hadi and Ives, Richard and Hasson, Yana and Noland, Eric and Cao, Yuan and Byrd, Nathan and Hou, Le and Wang, Qingze and Sottiaux, Thibault and Paganini, Michela and Lespiau, Jean-Baptiste and Moufarek, Alexandre and Hassan, Samer and Shivakumar, Kaushik and Amersfoort, Joost van and Mandhane, Amol and Joshi, Pratik and Goyal, Anirudh and Tung, Matthew and Brock, Andrew and Sheahan, Hannah and Misra, Vedant and Li, Cheng and Rakićević, Nemanja and Dehghani, Mostafa and Liu, Fangyu and Mittal, Sid and Oh, Junhyuk and Noury, Seb and Sezener, Eren and Huot, Fantine and Lamm, Matthew and Cao, Nicola De and Chen, Charlie and Mudgal, Sidharth and Stella, Romina and Brooks, Kevin and Vasudevan, Gautam and Liu, Chenxi and Chain, Mainak and Melinkeri, Nivedita and Cohen, Aaron and Wang, Venus and Seymore, Kristie and Zubkov, Sergey and Goel, Rahul and Yue, Summer and Krishnakumaran, Sai and Albert, Brian and Hurley, Nate and Sano, Motoki and Mohananey, Anhad and Joughin, Jonah and Filonov, Egor and Kępa, Tomasz and Eldawy, Yomna and Lim, Jiawern and Rishi, Rahul and Badiezadegan, Shirin and Bos, Taylor and Chang, Jerry and Jain, Sanil and Padmanabhan, Sri Gayatri Sundara and Puttagunta, Subha and Krishna, Kalpesh and Baker, Leslie and Kalb, Norbert and Bedapudi, Vamsi and Kurzrok, Adam and Lei, Shuntong and Yu, Anthony and Litvin, Oren and Zhou, Xiang and Wu, Zhichun and Sobell, Sam and Siciliano, Andrea and Papir, Alan and Neale, Robby and Bragagnolo, Jonas and Toor, Tej and Chen, Tina and Anklin, Valentin and Wang, Feiran and Feng, Richie and Gholami, Milad and Ling, Kevin and Liu, Lijuan and Walter, Jules and Moghaddam, Hamid and Kishore, Arun and Adamek, Jakub and Mercado, Tyler and Mallinson, Jonathan and Wandekar, Siddhinita and Cagle, Stephen and Ofek, Eran and Garrido, Guillermo and Lombriser, Clemens and Mukha, Maksim and Sun, Botu and Mohammad, Hafeezul Rahman and Matak, Josip and Qian, Yadi and Peswani, Vikas and Janus, Pawel and Yuan, Quan and Schelin, Leif and David, Oana and Garg, Ankur and He, Yifan and Duzhyi, Oleksii and Älgmyr, Anton and Lottaz, Timothée and Li, Qi and Yadav, Vikas and Xu, Luyao and Chinien, Alex and Shivanna, Rakesh and Chuklin, Aleksandr and Li, Josie and Spadine, Carrie and Wolfe, Travis and Mohamed, Kareem and Das, Subhabrata and Dai, Zihang and He, Kyle and Dincklage, Daniel von and Upadhyay, Shyam and Maurya, Akanksha and Chi, Luyan and Krause, Sebastian and Salama, Khalid and Rabinovitch, Pam G. and M, Pavan Kumar Reddy and Selvan, Aarush and Dektiarev, Mikhail and Ghiasi, Golnaz and Guven, Erdem and Gupta, Himanshu and Liu, Boyi and Sharma, Deepak and Shtacher, Idan Heimlich and Paul, Shachi and Akerlund, Oscar and Aubet, François-Xavier and Huang, Terry and Zhu, Chen and Zhu, Eric and Teixeira, Elico and Fritze, Matthew and Bertolini, Francesco and Marinescu, Liana-Eleonora and Bölle, Martin and Paulus, Dominik and Gupta, Khyatti and Latkar, Tejasi and Chang, Max and Sanders, Jason and Wilson, Roopa and Wu, Xuewei and Tan, Yi-Xuan and Thiet, Lam Nguyen and Doshi, Tulsee and Lall, Sid and Mishra, Swaroop and Chen, Wanming and Luong, Thang and Benjamin, Seth and Lee, Jasmine and Andrejczuk, Ewa and Rabiej, Dominik and Ranjan, Vipul and Styrc, Krzysztof and Yin, Pengcheng and Simon, Jon and Harriott, Malcolm Rose and Bansal, Mudit and Robsky, Alexei and Bacon, Geoff and Greene, David and Mirylenka, Daniil and Zhou, Chen and Sarvana, Obaid and Goyal, Abhimanyu and Andermatt, Samuel and Siegler, Patrick and Horn, Ben and Israel, Assaf and Pongetti, Francesco and Chen, Chih-Wei "Louis" and Selvatici, Marco and Silva, Pedro and Wang, Kathie and Tolins, Jackson and Guu, Kelvin and Yogev, Roey and Cai, Xiaochen and Agostini, Alessandro and Shah, Maulik and Nguyen, Hung and Donnaile, Noah Ó and Pereira, Sébastien and Friso, Linda and Stambler, Adam and Kurzrok, Adam and Kuang, Chenkai and Romanikhin, Yan and Geller, Mark and Yan, Z. J. and Jang, Kane and Lee, Cheng-Chun and Fica, Wojciech and Malmi, Eric and Tan, Qijun and Banica, Dan and Balle, Daniel and Pham, Ryan and Huang, Yanping and Avram, Diana and Shi, Hongzhi and Singh, Jasjot and Hidey, Chris and Ahuja, Niharika and Saxena, Pranab and Dooley, Dan and Potharaju, Srividya Pranavi and O'Neill, Eileen and Gokulchandran, Anand and Foley, Ryan and Zhao, Kai and Dusenberry, Mike and Liu, Yuan and Mehta, Pulkit and Kotikalapudi, Ragha and Safranek-Shrader, Chalence and Goodman, Andrew and Kessinger, Joshua and Globen, Eran and Kolhar, Prateek and Gorgolewski, Chris and Ibrahim, Ali and Song, Yang and Eichenbaum, Ali and Brovelli, Thomas and Potluri, Sahitya and Lahoti, Preethi and Baetu, Cip and Ghorbani, Ali and Chen, Charles and Crawford, Andy and Pal, Shalini and Sridhar, Mukund and Gurita, Petru and Mujika, Asier and Petrovski, Igor and Cedoz, Pierre-Louis and Li, Chenmei and Chen, Shiyuan and Santo, Niccolò Dal and Goyal, Siddharth and Punjabi, Jitesh and Kappaganthu, Karthik and Kwak, Chester and {LV}, Pallavi and Velury, Sarmishta and Choudhury, Himadri and Hall, Jamie and Shah, Premal and Figueira, Ricardo and Thomas, Matt and Lu, Minjie and Zhou, Ting and Kumar, Chintu and Jurdi, Thomas and Chikkerur, Sharat and Ma, Yenai and Yu, Adams and Kwak, Soo and Ähdel, Victor and Rajayogam, Sujeevan and Choma, Travis and Liu, Fei and Barua, Aditya and Ji, Colin and Park, Ji Ho and Hellendoorn, Vincent and Bailey, Alex and Bilal, Taylan and Zhou, Huanjie and Khatir, Mehrdad and Sutton, Charles and Rzadkowski, Wojciech and Macintosh, Fiona and Vij, Roopali and Shagin, Konstantin and Medina, Paul and Liang, Chen and Zhou, Jinjing and Shah, Pararth and Bi, Yingying and Dankovics, Attila and Banga, Shipra and Lehmann, Sabine and Bredesen, Marissa and Lin, Zifan and Hoffmann, John Eric and Lai, Jonathan and Chung, Raynald and Yang, Kai and Balani, Nihal and Bražinskas, Arthur and Sozanschi, Andrei and Hayes, Matthew and Alcalde, Héctor Fernández and Makarov, Peter and Chen, Will and Stella, Antonio and Snijders, Liselotte and Mandl, Michael and Kärrman, Ante and Nowak, Paweł and Wu, Xinyi and Dyck, Alex and Vaidyanathan, Krishnan and R, Raghavender and Mallet, Jessica and Rudominer, Mitch and Johnston, Eric and Mittal, Sushil and Udathu, Akhil and Christensen, Janara and Verma, Vishal and Irving, Zach and Santucci, Andreas and Elsayed, Gamaleldin and Davoodi, Elnaz and Georgiev, Marin and Tenney, Ian and Hua, Nan and Cideron, Geoffrey and Leurent, Edouard and Alnahlawi, Mahmoud and Georgescu, Ionut and Wei, Nan and Zheng, Ivy and Scandinaro, Dylan and Jiang, Heinrich and Snoek, Jasper and Sundararajan, Mukund and Wang, Xuezhi and Ontiveros, Zack and Karo, Itay and Cole, Jeremy and Rajashekhar, Vinu and Tumeh, Lara and Ben-David, Eyal and Jain, Rishub and Uesato, Jonathan and Datta, Romina and Bunyan, Oskar and Wu, Shimu and Zhang, John and Stanczyk, Piotr and Zhang, Ye and Steiner, David and Naskar, Subhajit and Azzam, Michael and Johnson, Matthew and Paszke, Adam and Chiu, Chung-Cheng and Elias, Jaume Sanchez and Mohiuddin, Afroz and Muhammad, Faizan and Miao, Jin and Lee, Andrew and Vieillard, Nino and Park, Jane and Zhang, Jiageng and Stanway, Jeff and Garmon, Drew and Karmarkar, Abhijit and Dong, Zhe and Lee, Jong and Kumar, Aviral and Zhou, Luowei and Evens, Jonathan and Isaac, William and Irving, Geoffrey and Loper, Edward and Fink, Michael and Arkatkar, Isha and Chen, Nanxin and Shafran, Izhak and Petrychenko, Ivan and Chen, Zhe and Jia, Johnson and Levskaya, Anselm and Zhu, Zhenkai and Grabowski, Peter and Mao, Yu and Magni, Alberto and Yao, Kaisheng and Snaider, Javier and Casagrande, Norman and Palmer, Evan and Suganthan, Paul and Castaño, Alfonso and Giannoumis, Irene and Kim, Wooyeol and Rybiński, Mikołaj and Sreevatsa, Ashwin and Prendki, Jennifer and Soergel, David and Goedeckemeyer, Adrian and Gierke, Willi and Jafari, Mohsen and Gaba, Meenu and Wiesner, Jeremy and Wright, Diana Gage and Wei, Yawen and Vashisht, Harsha and Kulizhskaya, Yana and Hoover, Jay and Le, Maigo and Li, Lu and Iwuanyanwu, Chimezie and Liu, Lu and Ramirez, Kevin and Khorlin, Andrey and Cui, Albert and {LIN}, Tian and Wu, Marcus and Aguilar, Ricardo and Pallo, Keith and Chakladar, Abhishek and Perng, Ginger and Abellan, Elena Allica and Zhang, Mingyang and Dasgupta, Ishita and Kushman, Nate and Penchev, Ivo and Repina, Alena and Wu, Xihui and Weide, Tom van der and Ponnapalli, Priya and Kaplan, Caroline and Simsa, Jiri and Li, Shuangfeng and Dousse, Olivier and Yang, Fan and Piper, Jeff and Ie, Nathan and Pasumarthi, Rama and Lintz, Nathan and Vijayakumar, Anitha and Andor, Daniel and Valenzuela, Pedro and Lui, Minnie and Paduraru, Cosmin and Peng, Daiyi and Lee, Katherine and Zhang, Shuyuan and Greene, Somer and Nguyen, Duc Dung and Kurylowicz, Paula and Hardin, Cassidy and Dixon, Lucas and Janzer, Lili and Choo, Kiam and Feng, Ziqiang and Zhang, Biao and Singhal, Achintya and Du, Dayou and {McKinnon}, Dan and Antropova, Natasha and Bolukbasi, Tolga and Keller, Orgad and Reid, David and Finchelstein, Daniel and Raad, Maria Abi and Crocker, Remi and Hawkins, Peter and Dadashi, Robert and Gaffney, Colin and Franko, Ken and Bulanova, Anna and Leblond, Rémi and Chung, Shirley and Askham, Harry and Cobo, Luis C. and Xu, Kelvin and Fischer, Felix and Xu, Jun and Sorokin, Christina and Alberti, Chris and Lin, Chu-Cheng and Evans, Colin and Dimitriev, Alek and Forbes, Hannah and Banarse, Dylan and Tung, Zora and Omernick, Mark and Bishop, Colton and Sterneck, Rachel and Jain, Rohan and Xia, Jiawei and Amid, Ehsan and Piccinno, Francesco and Wang, Xingyu and Banzal, Praseem and Mankowitz, Daniel J. and Polozov, Alex and Krakovna, Victoria and Brown, Sasha and Bateni, {MohammadHossein} and Duan, Dennis and Firoiu, Vlad and Thotakuri, Meghana and Natan, Tom and Geist, Matthieu and Girgin, Ser tan and Li, Hui and Ye, Jiayu and Roval, Ofir and Tojo, Reiko and Kwong, Michael and Lee-Thorp, James and Yew, Christopher and Sinopalnikov, Danila and Ramos, Sabela and Mellor, John and Sharma, Abhishek and Wu, Kathy and Miller, David and Sonnerat, Nicolas and Vnukov, Denis and Greig, Rory and Beattie, Jennifer and Caveness, Emily and Bai, Libin and Eisenschlos, Julian and Korchemniy, Alex and Tsai, Tomy and Jasarevic, Mimi and Kong, Weize and Dao, Phuong and Zheng, Zeyu and Liu, Frederick and Yang, Fan and Zhu, Rui and Teh, Tian Huey and Sanmiya, Jason and Gladchenko, Evgeny and Trdin, Nejc and Toyama, Daniel and Rosen, Evan and Tavakkol, Sasan and Xue, Linting and Elkind, Chen and Woodman, Oliver and Carpenter, John and Papamakarios, George and Kemp, Rupert and Kafle, Sushant and Grunina, Tanya and Sinha, Rishika and Talbert, Alice and Wu, Diane and Owusu-Afriyie, Denese and Du, Cosmo and Thornton, Chloe and Pont-Tuset, Jordi and Narayana, Pradyumna and Li, Jing and Fatehi, Saaber and Wieting, John and Ajmeri, Omar and Uria, Benigno and Ko, Yeongil and Knight, Laura and Héliou, Amélie and Niu, Ning and Gu, Shane and Pang, Chenxi and Li, Yeqing and Levine, Nir and Stolovich, Ariel and Santamaria-Fernandez, Rebeca and Goenka, Sonam and Yustalim, Wenny and Strudel, Robin and Elqursh, Ali and Deck, Charlie and Lee, Hyo and Li, Zonglin and Levin, Kyle and Hoffmann, Raphael and Holtmann-Rice, Dan and Bachem, Olivier and Arora, Sho and Koh, Christy and Yeganeh, Soheil Hassas and Põder, Siim and Tariq, Mukarram and Sun, Yanhua and Ionita, Lucian and Seyedhosseini, Mojtaba and Tafti, Pouya and Liu, Zhiyu and Gulati, Anmol and Liu, Jasmine and Ye, Xinyu and Chrzaszcz, Bart and Wang, Lily and Sethi, Nikhil and Li, Tianrun and Brown, Ben and Singh, Shreya and Fan, Wei and Parisi, Aaron and Stanton, Joe and Koverkathu, Vinod and Choquette-Choo, Christopher A. and Li, Yunjie and Lu, T. J. and Ittycheriah, Abe and Shroff, Prakash and Varadarajan, Mani and Bahargam, Sanaz and Willoughby, Rob and Gaddy, David and Desjardins, Guillaume and Cornero, Marco and Robenek, Brona and Mittal, Bhavishya and Albrecht, Ben and Shenoy, Ashish and Moiseev, Fedor and Jacobsson, Henrik and Ghaffarkhah, Alireza and Rivière, Morgane and Walton, Alanna and Crepy, Clément and Parrish, Alicia and Zhou, Zongwei and Farabet, Clement and Radebaugh, Carey and Srinivasan, Praveen and Salm, Claudia van der and Fidjeland, Andreas and Scellato, Salvatore and Latorre-Chimoto, Eri and Klimczak-Plucińska, Hanna and Bridson, David and Cesare, Dario de and Hudson, Tom and Mendolicchio, Piermaria and Walker, Lexi and Morris, Alex and Mauger, Matthew and Guseynov, Alexey and Reid, Alison and Odoom, Seth and Loher, Lucia and Cotruta, Victor and Yenugula, Madhavi and Grewe, Dominik and Petrushkina, Anastasia and Duerig, Tom and Sanchez, Antonio and Yadlowsky, Steve and Shen, Amy and Globerson, Amir and Webb, Lynette and Dua, Sahil and Li, Dong and Bhupatiraju, Surya and Hurt, Dan and Qureshi, Haroon and Agarwal, Ananth and Shani, Tomer and Eyal, Matan and Khare, Anuj and Belle, Shreyas Rammohan and Wang, Lei and Tekur, Chetan and Kale, Mihir Sanjay and Wei, Jinliang and Sang, Ruoxin and Saeta, Brennan and Liechty, Tyler and Sun, Yi and Zhao, Yao and Lee, Stephan and Nayak, Pandu and Fritz, Doug and Vuyyuru, Manish Reddy and Aslanides, John and Vyas, Nidhi and Wicke, Martin and Ma, Xiao and Eltyshev, Evgenii and Martin, Nina and Cate, Hardie and Manyika, James and Amiri, Keyvan and Kim, Yelin and Xiong, Xi and Kang, Kai and Luisier, Florian and Tripuraneni, Nilesh and Madras, David and Guo, Mandy and Waters, Austin and Wang, Oliver and Ainslie, Joshua and Baldridge, Jason and Zhang, Han and Pruthi, Garima and Bauer, Jakob and Yang, Feng and Mansour, Riham and Gelman, Jason and Xu, Yang and Polovets, George and Liu, Ji and Cai, Honglong and Chen, Warren and Sheng, {XiangHai} and Xue, Emily and Ozair, Sherjil and Angermueller, Christof and Li, Xiaowei and Sinha, Anoop and Wang, Weiren and Wiesinger, Julia and Koukoumidis, Emmanouil and Tian, Yuan and Iyer, Anand and Gurumurthy, Madhu and Goldenson, Mark and Shah, Parashar and Blake, M. K. and Yu, Hongkun and Urbanowicz, Anthony and Palomaki, Jennimaria and Fernando, Chrisantha and Durden, Ken and Mehta, Harsh and Momchev, Nikola and Rahimtoroghi, Elahe and Georgaki, Maria and Raul, Amit and Ruder, Sebastian and Redshaw, Morgan and Lee, Jinhyuk and Zhou, Denny and Jalan, Komal and Li, Dinghua and Hechtman, Blake and Schuh, Parker and Nasr, Milad and Milan, Kieran and Mikulik, Vladimir and Franco, Juliana and Green, Tim and Nguyen, Nam and Kelley, Joe and Mahendru, Aroma and Hu, Andrea and Howland, Joshua and Vargas, Ben and Hui, Jeffrey and Bansal, Kshitij and Rao, Vikram and Ghiya, Rakesh and Wang, Emma and Ye, Ke and Sarr, Jean Michel and Preston, Melanie Moranski and Elish, Madeleine and Li, Steve and Kaku, Aakash and Gupta, Jigar and Pasupat, Ice and Juan, Da-Cheng and Someswar, Milan and M, Tejvi and Chen, Xinyun and Amini, Aida and Fabrikant, Alex and Chu, Eric and Dong, Xuanyi and Muthal, Amruta and Buthpitiya, Senaka and Jauhari, Sarthak and Hua, Nan and Khandelwal, Urvashi and Hitron, Ayal and Ren, Jie and Rinaldi, Larissa and Drath, Shahar and Dabush, Avigail and Jiang, Nan-Jiang and Godhia, Harshal and Sachs, Uli and Chen, Anthony and Fan, Yicheng and Taitelbaum, Hagai and Noga, Hila and Dai, Zhuyun and Wang, James and Liang, Chen and Hamer, Jenny and Ferng, Chun-Sung and Elkind, Chenel and Atias, Aviel and Lee, Paulina and Listík, Vít and Carlen, Mathias and Kerkhof, Jan van de and Pikus, Marcin and Zaher, Krunoslav and Müller, Paul and Zykova, Sasha and Stefanec, Richard and Gatsko, Vitaly and Hirnschall, Christoph and Sethi, Ashwin and Xu, Xingyu Federico and Ahuja, Chetan and Tsai, Beth and Stefanoiu, Anca and Feng, Bo and Dhandhania, Keshav and Katyal, Manish and Gupta, Akshay and Parulekar, Atharva and Pitta, Divya and Zhao, Jing and Bhatia, Vivaan and Bhavnani, Yashodha and Alhadlaq, Omar and Li, Xiaolin and Danenberg, Peter and Tu, Dennis and Pine, Alex and Filippova, Vera and Ghosh, Abhipso and Limonchik, Ben and Urala, Bhargava and Lanka, Chaitanya Krishna and Clive, Derik and Sun, Yi and Li, Edward and Wu, Hao and Hongtongsak, Kevin and Li, Ianna and Thakkar, Kalind and Omarov, Kuanysh and Majmundar, Kushal and Alverson, Michael and Kucharski, Michael and Patel, Mohak and Jain, Mudit and Zabelin, Maksim and Pelagatti, Paolo and Kohli, Rohan and Kumar, Saurabh and Kim, Joseph and Sankar, Swetha and Shah, Vineet and Ramachandruni, Lakshmi and Zeng, Xiangkai and Bariach, Ben and Weidinger, Laura and Vu, Tu and Andreev, Alek and He, Antoine and Hui, Kevin and Kashem, Sheleem and Subramanya, Amar and Hsiao, Sissie and Hassabis, Demis and Kavukcuoglu, Koray and Sadovsky, Adam and Le, Quoc and Strohman, Trevor and Wu, Yonghui and Petrov, Slav and Dean, Jeffrey and Vinyals, Oriol},
	urldate = {2026-03-30},
	date = {2025-05-09},
	eprinttype = {arxiv},
	eprint = {2312.11805 [cs]},
}

@inproceedings{claude_nodate,
	title = {The Claude 3 Model Family: Opus, Sonnet, Haiku},
	url = {https://www.semanticscholar.org/paper/The-Claude-3-Model-Family:-Opus,-Sonnet,-Haiku/de8ba9b01c9ab7cbabf5c33b80b7bbc618857627},
	shorttitle = {The Claude 3 Model Family},
    author={Anthropic},
    date={2024},
	urldate = {2026-03-30},
}

@article{srivastava_beyond_2023,
	title = {Beyond the Imitation Game: Quantifying and extrapolating the capabilities of language models},
	issn = {2835-8856},
	url = {https://openreview.net/forum?id=uyTL5Bvosj&nesting=2&sort=date-desc},
	shorttitle = {Beyond the Imitation Game},
	journaltitle = {Transactions on Machine Learning Research},
	author = {Srivastava, Aarohi and Rastogi, Abhinav and Rao, Abhishek and Shoeb, Abu Awal Md and Abid, Abubakar and Fisch, Adam and Brown, Adam R. and Santoro, Adam and Gupta, Aditya and Garriga-Alonso, Adrià and Kluska, Agnieszka and Lewkowycz, Aitor and Agarwal, Akshat and Power, Alethea and Ray, Alex and Warstadt, Alex and Kocurek, Alexander W. and Safaya, Ali and Tazarv, Ali and Xiang, Alice and Parrish, Alicia and Nie, Allen and Hussain, Aman and Askell, Amanda and Dsouza, Amanda and Slone, Ambrose and Rahane, Ameet and Iyer, Anantharaman S. and Andreassen, Anders Johan and Madotto, Andrea and Santilli, Andrea and Stuhlmüller, Andreas and Dai, Andrew M. and La, Andrew and Lampinen, Andrew Kyle and Zou, Andy and Jiang, Angela and Chen, Angelica and Vuong, Anh and Gupta, Animesh and Gottardi, Anna and Norelli, Antonio and Venkatesh, Anu and Gholamidavoodi, Arash and Tabassum, Arfa and Menezes, Arul and Kirubarajan, Arun and Mullokandov, Asher and Sabharwal, Ashish and Herrick, Austin and Efrat, Avia and Erdem, Aykut and Karakaş, Ayla and Roberts, B. Ryan and Loe, Bao Sheng and Zoph, Barret and Bojanowski, Bartłomiej and Özyurt, Batuhan and Hedayatnia, Behnam and Neyshabur, Behnam and Inden, Benjamin and Stein, Benno and Ekmekci, Berk and Lin, Bill Yuchen and Howald, Blake and Orinion, Bryan and Diao, Cameron and Dour, Cameron and Stinson, Catherine and Argueta, Cedrick and Ferri, Cesar and Singh, Chandan and Rathkopf, Charles and Meng, Chenlin and Baral, Chitta and Wu, Chiyu and Callison-Burch, Chris and Waites, Christopher and Voigt, Christian and Manning, Christopher D. and Potts, Christopher and Ramirez, Cindy and Rivera, Clara E. and Siro, Clemencia and Raffel, Colin and Ashcraft, Courtney and Garbacea, Cristina and Sileo, Damien and Garrette, Dan and Hendrycks, Dan and Kilman, Dan and Roth, Dan and Freeman, C. Daniel and Khashabi, Daniel and Levy, Daniel and González, Daniel Moseguí and Perszyk, Danielle and Hernandez, Danny and Chen, Danqi and Ippolito, Daphne and Gilboa, Dar and Dohan, David and Drakard, David and Jurgens, David and Datta, Debajyoti and Ganguli, Deep and Emelin, Denis and Kleyko, Denis and Yuret, Deniz and Chen, Derek and Tam, Derek and Hupkes, Dieuwke and Misra, Diganta and Buzan, Dilyar and Mollo, Dimitri Coelho and Yang, Diyi and Lee, Dong-Ho and Schrader, Dylan and Shutova, Ekaterina and Cubuk, Ekin Dogus and Segal, Elad and Hagerman, Eleanor and Barnes, Elizabeth and Donoway, Elizabeth and Pavlick, Ellie and Rodolà, Emanuele and Lam, Emma and Chu, Eric and Tang, Eric and Erdem, Erkut and Chang, Ernie and Chi, Ethan A. and Dyer, Ethan and Jerzak, Ethan and Kim, Ethan and Manyasi, Eunice Engefu and Zheltonozhskii, Evgenii and Xia, Fanyue and Siar, Fatemeh and Martínez-Plumed, Fernando and Happé, Francesca and Chollet, Francois and Rong, Frieda and Mishra, Gaurav and Winata, Genta Indra and Melo, Gerard de and Kruszewski, Germàn and Parascandolo, Giambattista and Mariani, Giorgio and Wang, Gloria Xinyue and Jaimovitch-Lopez, Gonzalo and Betz, Gregor and Gur-Ari, Guy and Galijasevic, Hana and Kim, Hannah and Rashkin, Hannah and Hajishirzi, Hannaneh and Mehta, Harsh and Bogar, Hayden and Shevlin, Henry Francis Anthony and Schuetze, Hinrich and Yakura, Hiromu and Zhang, Hongming and Wong, Hugh Mee and Ng, Ian and Noble, Isaac and Jumelet, Jaap and Geissinger, Jack and Kernion, Jackson and Hilton, Jacob and Lee, Jaehoon and Fisac, Jaime Fernández and Simon, James B. and Koppel, James and Zheng, James and Zou, James and Kocon, Jan and Thompson, Jana and Wingfield, Janelle and Kaplan, Jared and Radom, Jarema and Sohl-Dickstein, Jascha and Phang, Jason and Wei, Jason and Yosinski, Jason and Novikova, Jekaterina and Bosscher, Jelle and Marsh, Jennifer and Kim, Jeremy and Taal, Jeroen and Engel, Jesse and Alabi, Jesujoba and Xu, Jiacheng and Song, Jiaming and Tang, Jillian and Waweru, Joan and Burden, John and Miller, John and Balis, John U. and Batchelder, Jonathan and Berant, Jonathan and Frohberg, Jörg and Rozen, Jos and Hernandez-Orallo, Jose and Boudeman, Joseph and Guerr, Joseph and Jones, Joseph and Tenenbaum, Joshua B. and Rule, Joshua S. and Chua, Joyce and Kanclerz, Kamil and Livescu, Karen and Krauth, Karl and Gopalakrishnan, Karthik and Ignatyeva, Katerina and Markert, Katja and Dhole, Kaustubh and Gimpel, Kevin and Omondi, Kevin and Mathewson, Kory Wallace and Chiafullo, Kristen and Shkaruta, Ksenia and Shridhar, Kumar and {McDonell}, Kyle and Richardson, Kyle and Reynolds, Laria and Gao, Leo and Zhang, Li and Dugan, Liam and Qin, Lianhui and Contreras-Ochando, Lidia and Morency, Louis-Philippe and Moschella, Luca and Lam, Lucas and Noble, Lucy and Schmidt, Ludwig and He, Luheng and Oliveros-Colón, Luis and Metz, Luke and Senel, Lütfi Kerem and Bosma, Maarten and Sap, Maarten and Hoeve, Maartje Ter and Farooqi, Maheen and Faruqui, Manaal and Mazeika, Mantas and Baturan, Marco and Marelli, Marco and Maru, Marco and Ramirez-Quintana, Maria Jose and Tolkiehn, Marie and Giulianelli, Mario and Lewis, Martha and Potthast, Martin and Leavitt, Matthew L. and Hagen, Matthias and Schubert, Mátyás and Baitemirova, Medina Orduna and Arnaud, Melody and {McElrath}, Melvin and Yee, Michael Andrew and Cohen, Michael and Gu, Michael and Ivanitskiy, Michael and Starritt, Michael and Strube, Michael and Swędrowski, Michał and Bevilacqua, Michele and Yasunaga, Michihiro and Kale, Mihir and Cain, Mike and Xu, Mimee and Suzgun, Mirac and Walker, Mitch and Tiwari, Mo and Bansal, Mohit and Aminnaseri, Moin and Geva, Mor and Gheini, Mozhdeh and T, Mukund Varma and Peng, Nanyun and Chi, Nathan Andrew and Lee, Nayeon and Krakover, Neta Gur-Ari and Cameron, Nicholas and Roberts, Nicholas and Doiron, Nick and Martinez, Nicole and Nangia, Nikita and Deckers, Niklas and Muennighoff, Niklas and Keskar, Nitish Shirish and Iyer, Niveditha S. and Constant, Noah and Fiedel, Noah and Wen, Nuan and Zhang, Oliver and Agha, Omar and Elbaghdadi, Omar and Levy, Omer and Evans, Owain and Casares, Pablo Antonio Moreno and Doshi, Parth and Fung, Pascale and Liang, Paul Pu and Vicol, Paul and Alipoormolabashi, Pegah and Liao, Peiyuan and Liang, Percy and Chang, Peter W. and Eckersley, Peter and Htut, Phu Mon and Hwang, Pinyu and Miłkowski, Piotr and Patil, Piyush and Pezeshkpour, Pouya and Oli, Priti and Mei, Qiaozhu and Lyu, Qing and Chen, Qinlang and Banjade, Rabin and Rudolph, Rachel Etta and Gabriel, Raefer and Habacker, Rahel and Risco, Ramon and Millière, Raphaël and Garg, Rhythm and Barnes, Richard and Saurous, Rif A. and Arakawa, Riku and Raymaekers, Robbe and Frank, Robert and Sikand, Rohan and Novak, Roman and Sitelew, Roman and Bras, Ronan Le and Liu, Rosanne and Jacobs, Rowan and Zhang, Rui and Salakhutdinov, Russ and Chi, Ryan Andrew and Lee, Seungjae Ryan and Stovall, Ryan and Teehan, Ryan and Yang, Rylan and Singh, Sahib and Mohammad, Saif M. and Anand, Sajant and Dillavou, Sam and Shleifer, Sam and Wiseman, Sam and Gruetter, Samuel and Bowman, Samuel R. and Schoenholz, Samuel Stern and Han, Sanghyun and Kwatra, Sanjeev and Rous, Sarah A. and Ghazarian, Sarik and Ghosh, Sayan and Casey, Sean and Bischoff, Sebastian and Gehrmann, Sebastian and Schuster, Sebastian and Sadeghi, Sepideh and Hamdan, Shadi and Zhou, Sharon and Srivastava, Shashank and Shi, Sherry and Singh, Shikhar and Asaadi, Shima and Gu, Shixiang Shane and Pachchigar, Shubh and Toshniwal, Shubham and Upadhyay, Shyam and Debnath, Shyamolima Shammie and Shakeri, Siamak and Thormeyer, Simon and Melzi, Simone and Reddy, Siva and Makini, Sneha Priscilla and Lee, Soo-Hwan and Torene, Spencer and Hatwar, Sriharsha and Dehaene, Stanislas and Divic, Stefan and Ermon, Stefano and Biderman, Stella and Lin, Stephanie and Prasad, Stephen and Piantadosi, Steven and Shieber, Stuart and Misherghi, Summer and Kiritchenko, Svetlana and Mishra, Swaroop and Linzen, Tal and Schuster, Tal and Li, Tao and Yu, Tao and Ali, Tariq and Hashimoto, Tatsunori and Wu, Te-Lin and Desbordes, Théo and Rothschild, Theodore and Phan, Thomas and Wang, Tianle and Nkinyili, Tiberius and Schick, Timo and Kornev, Timofei and Tunduny, Titus and Gerstenberg, Tobias and Chang, Trenton and Neeraj, Trishala and Khot, Tushar and Shultz, Tyler and Shaham, Uri and Misra, Vedant and Demberg, Vera and Nyamai, Victoria and Raunak, Vikas and Ramasesh, Vinay Venkatesh and Prabhu, Vinay Uday and Padmakumar, Vishakh and Srikumar, Vivek and Fedus, William and Saunders, William and Zhang, William and Vossen, Wout and Ren, Xiang and Tong, Xiaoyu and Zhao, Xinran and Wu, Xinyi and Shen, Xudong and Yaghoobzadeh, Yadollah and Lakretz, Yair and Song, Yangqiu and Bahri, Yasaman and Choi, Yejin and Yang, Yichi and Hao, Sophie and Chen, Yifu and Belinkov, Yonatan and Hou, Yu and Hou, Yufang and Bai, Yuntao and Seid, Zachary and Zhao, Zhuoye and Wang, Zijian and Wang, Zijie J. and Wang, Zirui and Wu, Ziyi},
	urldate = {2026-03-30},
	date = {2023-01-19},
	langid = {english},
}

@article{kung_performance_2023,
	title = {Performance of {ChatGPT} on {USMLE}: Potential for {AI}-assisted medical education using large language models},
	volume = {2},
	issn = {2767-3170},
	url = {https://journals.plos.org/digitalhealth/article?id=10.1371/journal.pdig.0000198},
	doi = {10.1371/journal.pdig.0000198},
	shorttitle = {Performance of {ChatGPT} on {USMLE}},
	pages = {e0000198},
	number = {2},
	journaltitle = {{PLOS} Digital Health},
	shortjournal = {{PLOS} Digital Health},
	publisher = {Public Library of Science},
	author = {Kung, Tiffany H. and Cheatham, Morgan and Medenilla, Arielle and Sillos, Czarina and Leon, Lorie De and Elepaño, Camille and Madriaga, Maria and Aggabao, Rimel and Diaz-Candido, Giezel and Maningo, James and Tseng, Victor},
	urldate = {2026-03-31},
	date = {2023-02-09},
	langid = {english},
}

@inproceedings{garcia_dataset_2020,
	location = {Cham},
	title = {A Dataset and Baselines for Visual Question Answering on Art},
	isbn = {978-3-030-66096-3},
	doi = {10.1007/978-3-030-66096-3_8},
	pages = {92--108},
	booktitle = {Computer Vision – {ECCV} 2020 Workshops},
	publisher = {Springer International Publishing},
	author = {Garcia, Noa and Ye, Chentao and Liu, Zihua and Hu, Qingtao and Otani, Mayu and Chu, Chenhui and Nakashima, Yuta and Mitamura, Teruko},
	editor = {Bartoli, Adrien and Fusiello, Andrea},
	date = {2020},
	langid = {english},
}

@misc{spinaci_benchmarking_2025,
	title = {Benchmarking Vision-Language and Multimodal Large Language Models in Zero-shot and Few-shot Scenarios: A study on Christian Iconography},
	url = {http://arxiv.org/abs/2509.18839},
	doi = {10.48550/arXiv.2509.18839},
	shorttitle = {Benchmarking Vision-Language and Multimodal Large Language Models in Zero-shot and Few-shot Scenarios},
	number = {{arXiv}:2509.18839},
	publisher = {{arXiv}},
	author = {Spinaci, Gianmarco and Klic, Lukas and Colavizza, Giovanni},
	urldate = {2026-03-31},
	date = {2025-09-23},
	eprinttype = {arxiv},
	eprint = {2509.18839 [cs]},
}

@article{becattini_viscounth_2023,
	title = {{VISCOUNTH}: A Large-scale Multilingual Visual Question Answering Dataset for Cultural Heritage},
	volume = {19},
	issn = {1551-6857},
	url = {https://dl.acm.org/doi/10.1145/3590773},
	doi = {10.1145/3590773},
	shorttitle = {{VISCOUNTH}},
	pages = {193:1--193:20},
	number = {6},
	journaltitle = {{ACM} Trans. Multimedia Comput. Commun. Appl.},
	author = {Becattini, Federico and Bongini, Pietro and Bulla, Luana and Bimbo, Alberto Del and Marinucci, Ludovica and Mongiovì, Misael and Presutti, Valentina},
	urldate = {2026-03-31},
	date = {2023},
}

@misc{romero_cvqa_2024,
	title = {{CVQA}: Culturally-diverse Multilingual Visual Question Answering Benchmark},
	url = {http://arxiv.org/abs/2406.05967},
	doi = {10.48550/arXiv.2406.05967},
	shorttitle = {{CVQA}},
	number = {{arXiv}:2406.05967},
	publisher = {{arXiv}},
	author = {Romero, David and Lyu, Chenyang and Wibowo, Haryo Akbarianto and Lynn, Teresa and Hamed, Injy and Kishore, Aditya Nanda and Mandal, Aishik and Dragonetti, Alina and Abzaliev, Artem and Tonja, Atnafu Lambebo and Balcha, Bontu Fufa and Whitehouse, Chenxi and Salamea, Christian and Velasco, Dan John and Adelani, David Ifeoluwa and Meur, David Le and Villa-Cueva, Emilio and Koto, Fajri and Farooqui, Fauzan and Belcavello, Frederico and Batnasan, Ganzorig and Vallejo, Gisela and Caulfield, Grainne and Ivetta, Guido and Song, Haiyue and Ademtew, Henok Biadglign and Maina, Hernán and Lovenia, Holy and Azime, Israel Abebe and Cruz, Jan Christian Blaise and Gala, Jay and Geng, Jiahui and Ortiz-Barajas, Jesus-German and Baek, Jinheon and Dunstan, Jocelyn and Alemany, Laura Alonso and Nagasinghe, Kumaranage Ravindu Yasas and Benotti, Luciana and D'Haro, Luis Fernando and Viridiano, Marcelo and Estecha-Garitagoitia, Marcos and Cabrera, Maria Camila Buitrago and Rodríguez-Cantelar, Mario and Jouitteau, Mélanie and Mihaylov, Mihail and Imam, Mohamed Fazli Mohamed and Adilazuarda, Muhammad Farid and Gochoo, Munkhjargal and Otgonbold, Munkh-Erdene and Etori, Naome and Niyomugisha, Olivier and Silva, Paula Mónica and Chitale, Pranjal and Dabre, Raj and Chevi, Rendi and Zhang, Ruochen and Diandaru, Ryandito and Cahyawijaya, Samuel and Góngora, Santiago and Jeong, Soyeong and Purkayastha, Sukannya and Kuribayashi, Tatsuki and Clifford, Teresa and Jayakumar, Thanmay and Torrent, Tiago Timponi and Ehsan, Toqeer and Araujo, Vladimir and Kementchedjhieva, Yova and Burzo, Zara and Lim, Zheng Wei and Yong, Zheng Xin and Ignat, Oana and Nwatu, Joan and Mihalcea, Rada and Solorio, Thamar and Aji, Alham Fikri},
	urldate = {2026-03-31},
	date = {2024-11-04},
	eprinttype = {arxiv},
	eprint = {2406.05967 [cs]},
}

@inproceedings{goyal_making_2017,
	title = {Making the v in {VQA} Matter: Elevating the Role of Image Understanding in Visual Question Answering},
	url = {https://openaccess.thecvf.com/content_cvpr_2017/html/Goyal_Making_the_v_CVPR_2017_paper.html},
	shorttitle = {Making the v in {VQA} Matter},
	eventtitle = {Proceedings of the {IEEE} Conference on Computer Vision and Pattern Recognition},
	pages = {6904--6913},
	author = {Goyal, Yash and Khot, Tejas and Summers-Stay, Douglas and Batra, Dhruv and Parikh, Devi},
	urldate = {2026-03-31},
	date = {2017},
}

@article{katz_gpt-4_2024,
	title = {{GPT}-4 passes the bar exam},
	volume = {382},
	issn = {1471-2962},
	doi = {10.1098/rsta.2023.0254},
	pages = {20230254},
	number = {2270},
	journaltitle = {Philosophical Transactions. Series A, Mathematical, Physical, and Engineering Sciences},
	shortjournal = {Philos Trans A Math Phys Eng Sci},
	author = {Katz, Daniel Martin and Bommarito, Michael James and Gao, Shang and Arredondo, Pablo},
	date = {2024-04-15},
}

@inproceedings{johnson_clevr_2017,
	title = {{CLEVR}: A Diagnostic Dataset for Compositional Language and Elementary Visual Reasoning},
	url = {https://openaccess.thecvf.com/content_cvpr_2017/html/Johnson_CLEVR_A_Diagnostic_CVPR_2017_paper.html},
	shorttitle = {{CLEVR}},
	eventtitle = {Proceedings of the {IEEE} Conference on Computer Vision and Pattern Recognition},
	pages = {2901--2910},
	author = {Johnson, Justin and Hariharan, Bharath and van der Maaten, Laurens and Fei-Fei, Li and Lawrence Zitnick, C. and Girshick, Ross},
	urldate = {2026-05-13},
	date = {2017},
}

@misc{zhou_lost_2026,
	title = {Lost in Benchmarks? Rethinking Large Language Model Benchmarking with Item Response Theory},
	url = {http://arxiv.org/abs/2505.15055},
	doi = {10.48550/arXiv.2505.15055},
	shorttitle = {Lost in Benchmarks?},
	number = {{arXiv}:2505.15055},
	publisher = {{arXiv}},
	author = {Zhou, Hongli and Huang, Hui and Zhao, Ziqing and Han, Lvyuan and Wang, Huicheng and Chen, Kehai and Yang, Muyun and Bao, Wei and Dong, Jian and Xu, Bing and Zhu, Conghui and Cao, Hailong and Zhao, Tiejun},
	urldate = {2026-05-24},
	date = {2026-01-16},
	eprinttype = {arxiv},
	eprint = {2505.15055 [cs.CL]},
}

@misc{polo_tinybenchmarks_2024,
	title = {{tinyBenchmarks}: evaluating {LLMs} with fewer examples},
	url = {http://arxiv.org/abs/2402.14992},
	doi = {10.48550/arXiv.2402.14992},
	shorttitle = {{tinyBenchmarks}},
	number = {{arXiv}:2402.14992},
	publisher = {{arXiv}},
	author = {Polo, Felipe Maia and Weber, Lucas and Choshen, Leshem and Sun, Yuekai and Xu, Gongjun and Yurochkin, Mikhail},
	urldate = {2026-03-23},
	date = {2024-02-22},
	eprinttype = {arxiv},
	eprint = {2402.14992 [cs]},
	note = {version: 1},
}

@inproceedings{tong_eyes_2024,
	title = {Eyes Wide Shut? Exploring the Visual Shortcomings of Multimodal {LLMs}},
	url = {https://openaccess.thecvf.com/content/CVPR2024/html/Tong_Eyes_Wide_Shut_Exploring_the_Visual_Shortcomings_of_Multimodal_LLMs_CVPR_2024_paper.html},
	shorttitle = {Eyes Wide Shut?},
	eventtitle = {Proceedings of the {IEEE}/{CVF} Conference on Computer Vision and Pattern Recognition},
	pages = {9568--9578},
	author = {Tong, Shengbang and Liu, Zhuang and Zhai, Yuexiang and Ma, Yi and {LeCun}, Yann and Xie, Saining},
	urldate = {2026-05-13},
	date = {2024},
	langid = {english},
}

@inproceedings{zhang_mathverse_2025,
	location = {Cham},
	title = {{MATHVERSE}: Does Your Multi-modal {LLM} Truly See the Diagrams in Visual Math Problems?},
	isbn = {978-3-031-73242-3},
	doi = {10.1007/978-3-031-73242-3_10},
	shorttitle = {{MATHVERSE}},
	pages = {169--186},
	booktitle = {Computer Vision – {ECCV} 2024},
	publisher = {Springer Nature Switzerland},
	author = {Zhang, Renrui and Jiang, Dongzhi and Zhang, Yichi and Lin, Haokun and Guo, Ziyu and Qiu, Pengshuo and Zhou, Aojun and Lu, Pan and Chang, Kai-Wei and Qiao, Yu and Gao, Peng and Li, Hongsheng},
	editor = {Leonardis, Aleš and Ricci, Elisa and Roth, Stefan and Russakovsky, Olga and Sattler, Torsten and Varol, Gül},
	date = {2025},
	langid = {english},
}

@inproceedings{zellers_recognition_2019,
	title = {From Recognition to Cognition: Visual Commonsense Reasoning},
	url = {https://openaccess.thecvf.com/content_CVPR_2019/html/Zellers_From_Recognition_to_Cognition_Visual_Commonsense_Reasoning_CVPR_2019_paper.html},
	shorttitle = {From Recognition to Cognition},
	eventtitle = {Proceedings of the {IEEE}/{CVF} Conference on Computer Vision and Pattern Recognition},
	pages = {6720--6731},
	author = {Zellers, Rowan and Bisk, Yonatan and Farhadi, Ali and Choi, Yejin},
	urldate = {2026-05-13},
	date = {2019},
}

@article{siam_benchmarking_2025,
	title = {Benchmarking large language models on the United States medical licensing examination for clinical reasoning and medical licensing scenarios},
	volume = {16},
	rights = {2025 The Author(s)},
	issn = {2045-2322},
	url = {https://www.nature.com/articles/s41598-025-31010-4},
	doi = {10.1038/s41598-025-31010-4},
	pages = {1387},
	number = {1},
	journaltitle = {Scientific Reports},
	shortjournal = {Sci Rep},
	publisher = {Nature Publishing Group},
	author = {Siam, Md Kamrul and Varela, Angel and Faruk, Md Jobair Hossain and Cheng, Jerry Q. and Gu, Huanying and Maruf, Abdullah Al and Aung, Zeyar},
	urldate = {2026-03-31},
	date = {2025-12-03},
	langid = {english},
}





% \end{thebibliography}

\appendix

\section{Glossary} \label{ap:glossary}

\textbf{Accuracy.} The proportion of items a model answers correctly under a chosen scoring rule. EduArt uses different scoring rules per format (exact match for MCQ radio; F1 for MCQ check and select errors; statement-level accuracy for true/false; element accuracy for positioning; blank-level accuracy for completion).

\textbf{Benchmark.} A standardised set of inputs (here, exam questions) with known correct answers, used to compare the performance of different models on a defined task.

\textbf{Classical Test Theory (CTT).} A psychometric framework that describes each test item by its difficulty (the proportion of respondents who answer it correctly) and its discrimination (how strongly performance on the item correlates with total test score).

\textbf{Confidence interval (CI).} A range of values that, with a stated probability (here 95 percent), is expected to contain the true value of a quantity estimated from data.

\textbf{Discrimination ($r_{pb}$)}. The point-biserial correlation between scoring correctly on an item and the model's total score across the benchmark. Higher values mean the item separates stronger models from weaker ones.

\textbf{Exact match.} A binary scoring rule that awards full credit only when the model's output matches the ground-truth answer character-for-character (after normalisation).

\textbf{F1 score.} A scoring metric defined as the harmonic mean of precision (proportion of selected options that are correct) and recall (proportion of correct options that are selected). Used here for formats with multiple correct answers per item.

\textbf{Item Response Theory (IRT).} A psychometric framework that models the probability of a correct response as a function of both the respondent's ability and item-level parameters (difficulty, discrimination, and sometimes guessing).

\textbf{Logistic regression.} A statistical model that estimates the probability of a binary outcome (here, a correct answer) as a function of predictors. Coefficients are reported as odds ratios.

\textbf{Macro-average.} An average computed by first calculating accuracy within each category (here, each question format) and then averaging across categories, giving equal weight to each category regardless of size.

\textbf{Multimodal large language model.} A model that accepts both text and other modalities (typically images) as input and produces text as output.

\textbf{Odds ratio (OR).} The ratio of the odds of an outcome under one condition to the odds under a reference condition. OR > 1 means the predictor is associated with higher probability of a correct answer; OR < 1 means lower probability.

\textbf{Saturation.} The condition in which a benchmark can no longer distinguish among the best-performing models because their accuracies are all near the maximum possible value.

\textbf{Visual Question Answering (VQA).} An evaluation paradigm in which a model receives an image and a natural-language question and produces a natural-language answer.

\textbf{Zero-shot evaluation.} Evaluation of a model on a task without any task-specific training or in-context examples, relying only on the prompt and the model's pre-trained knowledge.

\end{document}